\documentclass{article}

\PassOptionsToPackage{numbers, compress}{natbib}

\usepackage[preprint]{neurips_data_2024}

\usepackage[utf8]{inputenc} 
\usepackage[T1]{fontenc}    
\usepackage{hyperref}       
\usepackage{url}            
\usepackage{booktabs}       
\usepackage{amsfonts}       
\usepackage{nicefrac}       
\usepackage{microtype}      
\usepackage{array}
\usepackage{multicol}
\usepackage{multirow}
\usepackage{tabularx}
\usepackage[table]{xcolor}
\usepackage[T1]{fontenc}
\usepackage{graphicx}
\usepackage{enumitem}
\usepackage{amsmath}

\usepackage{makecell}

\newcommand{\customhline}{\noalign{\vspace{1.25pt}}\hline\noalign{\vspace{1.25pt}}}
\newcommand{\customtop}{\toprule\noalign{\vspace{1.25pt}}}
\newcommand{\custommid}{\noalign{\vspace{1.25pt}}\midrule\noalign{\vspace{1.25pt}}}
\newcommand{\custombot}{\noalign{\vspace{1.25pt}}\bottomrule}

\definecolor{bb}{rgb}{0.12,0.565,1}
\definecolor{gg}{rgb}{0.2,0.8,0.2}
\definecolor{rr}{rgb}{1,0.85,0.2}

\newif\ifdraft
\draftfalse

\newcommand{\ours}[0]{M3GIA}

\title{M3GIA: A Cognition Inspired Multilingual and Multimodal General Intelligence Ability Benchmark}

\author{
    Wei Song$^{1,2,3}$\thanks{Co-first authors. Work done during Wei Song's internship at Alibaba. }\enskip\enskip 
    Yadong Li$^{2}$\footnotemark[1]\enskip\enskip
    Jianhua Xu$^{2}$\thanks{Co-second authors. }\enskip\enskip
    Guowei Wu$^{4}$\footnotemark[2]\enskip\enskip
    Lingfeng Ming$^{2}$\enskip\enskip
    Kexin Yi$^{2}$\enskip\enskip \\
    {\bfseries Weihua Luo$^{2}$}\enskip\enskip
    {\bfseries Houyi Li$^{2}$}\enskip\enskip
    {\bfseries Yi Du$^{4}$}\enskip\enskip
    {\bfseries Fangda Guo$^{5}$}\enskip\enskip
    {\bfseries Kaicheng Yu$^{1}$}\thanks{Corresponding author. }\\
    \textsuperscript{1}AutoLab, Westlake University \enskip 
    \textsuperscript{2}AI Business, Alibaba Group \enskip
    \textsuperscript{3}Zhejiang University \enskip \\
    \textsuperscript{4}Key Laboratory of Behavioral Science, Institute of Psychology, CAS \enskip \\
    \textsuperscript{5}Key Laboratory of AI Safety, Institute of Computing Technology, CAS \enskip \\
    \texttt{\{songwei, kyu\}@westlake.edu.cn, adonlee.lyd@alibaba-inc.com}
}

\begin{document}

\maketitle

\begin{abstract}
    As recent multi-modality large language models~(MLLMs) have shown formidable proficiency on various complex tasks, there has been increasing attention on debating whether these models could eventually mirror human intelligence.
    However, existing benchmarks mainly focus on evaluating solely on task performance, such as the accuracy of identifying the attribute of an object. Combining well-developed cognitive science to understand the intelligence of MLLMs beyond superficial achievements remains largely unexplored. To this end, we introduce the first cognitive-driven multi-lingual and multi-modal benchmark to evaluate the general intelligence ability of MLLMs, dubbed \ours{}. Specifically, we identify five key cognitive factors based on the well-recognized Cattell-Horn-Carrol~(CHC) model of intelligence and propose a novel evaluation metric. In addition, since most MLLMs are trained to perform in different languages, a natural question arises: is language a key factor influencing the cognitive ability of MLLMs? As such, we go beyond English to encompass other languages based on their popularity, including Chinese, French, Spanish, Portuguese and Korean, to construct our \ours{}. We make sure all the data relevant to the cultural backgrounds are collected from their native context to avoid English-centric bias. 
    We collected a significant corpus of data from human participants, revealing that the most advanced MLLM reaches the lower boundary of human intelligence in English. Yet, there remains a pronounced disparity in the other five languages assessed. We also reveals an interesting \emph{winner takes all} phenomenon that are aligned with the discovery in cognitive studies. Our benchmark will be open-sourced, with the aspiration of facilitating the enhancement of cognitive capabilities in MLLMs.
    
\end{abstract}

\section{Introduction}
\label{intro}

In 1956, researchers across different domains, including mathematics, cognitive psychology and computer science, pointed out an interesting direction, dubbed artificial intelligence~(AI). The formal definition is \emph{``The study is to proceed on the basis of the conjecture that every aspect of learning or any other feature of intelligence can in principle be so precisely described that a machine can be made to simulate it.''~\cite{mccarthy2006proposal}}. 
Through extensive efforts in pursuing artificial intelligence, the field has converged to a paradigm of data-driven machine learning models, which are still deeply intertwined with cognitive science as they often mirror basic cognitive mechanisms, e.g. convolutional neural networks~\cite{krizhevsky2012imagenet} and the attention mechanism~\cite{vaswani2017attention}.
Recent advances, such as GPT-4o~\cite{GPT-4o}, demonstrate that these MLLMs can outperform  human on various complex tasks~\cite{achiam2023gpt,wang2023emotional} and shed light to emergent ability with the increasing scale of data and model size~\cite{wei2022emergent}. In light of these developments, our aim is to evaluate these state-of-the-art models through the lens of cognitive science, as it directly aligns with the primary motivation of AI research.

To explore the mental intelligence emerging from these large models, efforts have been directed toward analyzing these models from a psychological perspective. Some pioneering works report that LLMs have demonstrated human-like cognition~\cite{binz2023using,kosinski2023theory}. For instance, Theory of mind (ToM) has been applied to assess large models, revealing that GPT-4 exhibits ToM capabilities similar to human inference patterns~\cite{bubeck2023sparks,kosinski2023theory,gandhi2024understanding}.
Meanwhile, Multimodal Large Language Models (MLLMs), which use powerful LLMs as brain to process and integrate multimodal information, have exhibited impressive emergent abilities, such as generating website code from images~\cite{zhu2023minigpt}, understanding the meaning of a meme~\cite{yang2023mm}, and math reasoning~\cite{driess2023palm}.
Thanks to their ability to process information from a broader spectrum of sources, they exhibit a more holistic cognitive process, resembling human cognition more closely than models confined to purely linguistic input.

\begin{figure}
    \centering
    \vspace{-0.5cm}
    \includegraphics[width=\textwidth]{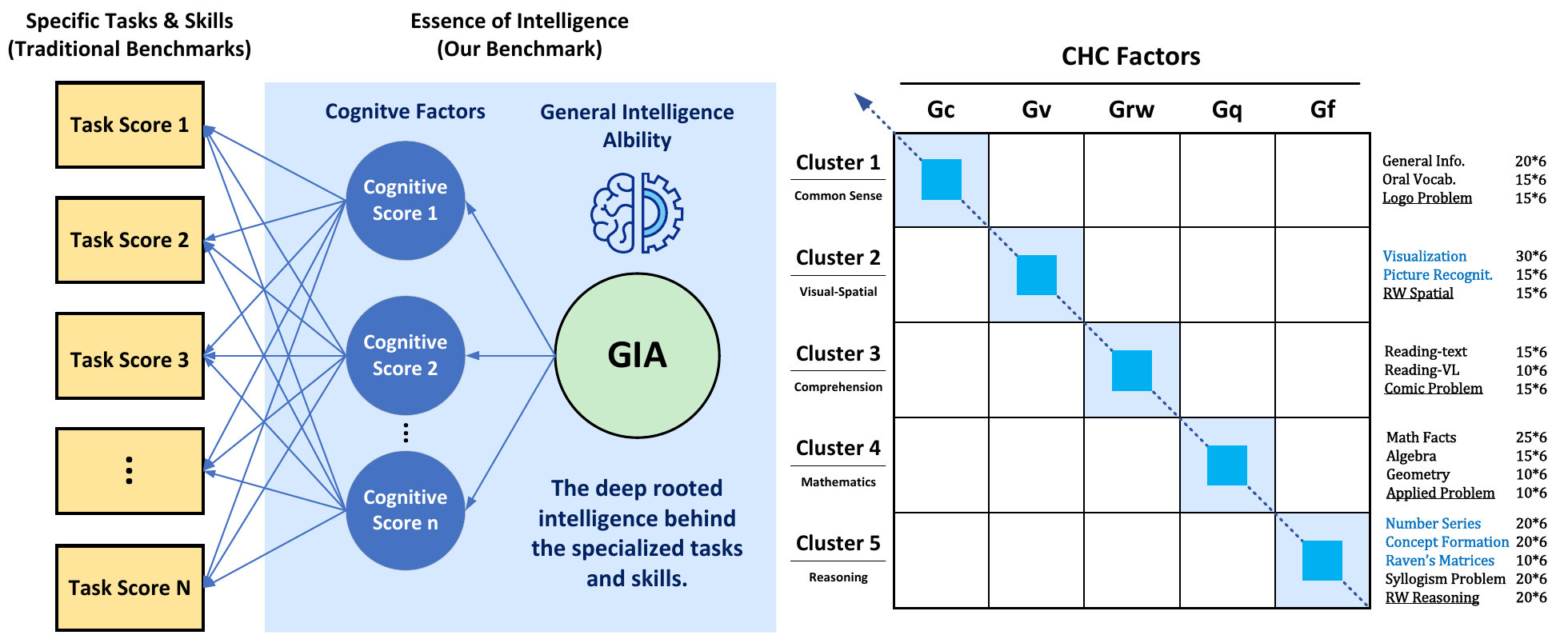}
    \caption{\textbf{Overview of multi-lingual multi-modal general intelligence ability benchmark.}
    \textbf{(Left)} In contrast to traditional benchmarks that focus on evaluating specific task performances, we draw inspiration from cognitive science to categorize five cognitive factors, try to provide a feasible evaluation of general intelligence ability~(GIA). \textbf{(Right)}  Specifically, we adopt the factors from the CHC theory to disentangle fundamental cognitive abilities with existing evaluation tasks. In addition, to further understand how language impacts such ability, we collect or design questions in six languages with large population. }
    \vspace{-0.3cm}
    \label{teaser}
\end{figure}

Existing multi-modality benchmarks, such as MMBench~\cite{liu2023mmbench}, MME~\cite{fu2024mme}, and MM-Vet~\cite{yu2023mm}, have made the attempt to compartmentalize model capabilities across multiple tasks.
For instance, MMBench covers 20 different abilities, encompassing function reasoning, physical property reasoning, object localization and social reasoning.
However, they often fail to provide a persuasive explanation for their selection of dimensions, as they tend to be mired in subjectivity and lack a solid theoretical underpinning. 
Moreover, as depicted in Figure~\ref{teaser}~(left), their ability dimensions are still rather task-oriented, neglecting a systematic evaluation of the models' underlying cognitive abilities that govern task performance through the lens of cognitive science. This oversight raises concerns that benchmarks might devolve into mere training targets rather than instruments for true insight, failing to provide a holistic measure of the models' capabilities~\cite{schaeffer2024emergent}.
In short, the ability to solve specific tasks is insufficient to reflect the true level of intelligence, as supported by a psychological study\cite{poldrack2016brain}, and formally evaluating the cognitive factors of MLLMs remains largely unexplored.

In this paper, we close the gap by introducing the first benchmark that comprehensively evaluate the cognitive abilities of MLLMs under the theoretical umbrella of the well-recognized Cattell-Horn-Carroll (CHC) Model of Intelligence~\cite{schneider2012cattell}, dubbed \ours{}. 
As in Figure~\ref{teaser}(right), based on the CHC Model, we categorizes the cognitive capacities of current MLLMs into five dimensions: Fluid reasoning (Gf), Comprehension-Knowledge (Gc), Visual processing (Gv), Reading and Writing (Grw), Quantitative knowledge (Gq), and collect corresponding questions as a measurement. In addition, as using multi-lingual data to scale up the capability of MLLMs becomes a de-facto standard, we are curious whether languages make any impact on their cognitive abilities. As such, we extend our benchmark to include five more languages, including Chinese, Spanish, French, Portuguese and Korean roughly based on their population, to disentangle the language factor with cognitive ability.

To evenly assess the five cognitive dimensions, we refer to human intelligence tests, such as Raven's Progressive Matrices Test~\cite{raven2003raven} and the Woodcock-Johnson IV Tests of Cognitive Abilities (WJ IV)~\cite{schrank2018woodcock}, and establish broad question types that correspond to the these cognitive dimensions, which are further subdivided into 18 narrow question types (see later Sec.~\ref{method}). All in all, our \ours{} contains 1,800 questions, where over half are carefully designed from scratch following the standard. The test for each language maintain consistency in terms of the number of questions, structure, and distribution of question types. In addition, to highlight the multilingual nature of our benchmark, we collect data relevant to cultural backgrounds from native language sources rather than simply translating them from English, thereby avoiding the English-centric bias. 

We evaluate 24 MLLMs, including the state-of-the-art close and open-sourced ones. In general, The latest advancements in MLLMs have achieved performance levels that fall within the lower boundary of human intelligence in English. Yet, there remains a pronounced disparity in the other five languages assessed.
We also notice that MLLMs' proficiency in one cognitive domain often translates into superior performance across other domains as well. This phenomenon interestingly aligns with the pattern observed in human intelligence which empirically suggests the existence of General Intelligence Ability (GIA) in MLLMs.

\section{Related works}
\label{related}

\paragraph{Evaluation Benchmark for MLLMs.} 
As multimodal large language models (MLLMs) exhibit remarkable generalization capabilities across a broad spectrum of downstream tasks, relying exclusively on their performance within single vision-language tasks — such as visual recognition~\cite{goyal2017making}, image description~\cite{chen2015microsoft,agrawal2019nocaps,young2014image}, scene text understanding~\cite{singh2019towards,sidorov2020textcaps}, and external knowledge~\cite{marino2019ok} — is insufficient to fully uncover the comprehensive performance of MLLMs.
People then turn to a new paradigm to construct all-round benchmarks to assess a broader spectrum of challenging multimodal tasks~\cite{yin2024lamm,xu2023lvlm,li2023seed,liu2023mmbench,fu2024mme,yu2023mm}.
Another trend in MLLM assessment is the use of human exam questions~\cite{lu2023mathvista,lu2022learn,zhong2023agieval,yue2023mmmu,zhang2024m3exam}. For instance, AGIEval~\cite{zhong2023agieval} sources questions from standardized exams such as college entrance exams and lawyer qualification tests. 
While these benchmarks makes progresses in evaluating the human-centric ability of MLLMs, it may not be suitable to evaluate the intelligence of MLLMs because research in psychological field points out that the superficial performance on tasks alone cannot be a solid indicator for human's intelligence.\cite{poldrack2016brain}

\paragraph{General Intelligence Ability and the CHC Theory.} 

Arising from the empirical fact that an individual's proficiency in one area frequently correlates with high performance in other areas, Charles Spearman first introduced General Intelligence Ability (GIA) in 1904~\cite{spearman1961general}.  This construct refers to the idea that a single underlying factor, known as the g-factor, can account for the positive correlations among cognitive abilities and reflect the general intelligence that fundamentally underlies an individual's intelligence.
To concretely understand GIA, numerous attempts has been made to model the structure of human cognition.
John Carroll’s Three-Stratum Model~\cite{Carroll_1993} elaborated on this with a hierarchical structure of intelligence, including a general ``g'' factor and specific cognitive abilities. Howard Gardner’s Multiple Intelligence Theory~\cite{flynn1987massive} proposed diverse forms of intelligence, while Sternberg’s Triarchic Theory~\cite{sternberg1985beyond} focused on practical, creative, and analytical aspects. 
These theories collectively contributed to the development of the Cattell-Horn-Carroll (CHC) model of intelligence, which is the most comprehensive and empirically validated structural model of human cognition~\cite{mcgrew2004internal} to date, integrating various aspects of cognition into a unified framework.
Recent study~\cite{coda2024cogbench}, has primarily focused on evaluating the performance of large language models on sophisticated psychological tasks, neglecting the assessment of models' intelligence from the foundational standpoint of cognitive models. Our \ours{} constitutes the first attempt to bring the latest cognitive science modeling of intelligence into MLLMs evaluation to address this gap.

\section{\ours{}}
\label{method}


Concretely, we introduce the first cognition inspired multi-linguistic and multi-modal benchmark to evaluate the general intelligence accuracy of large models. In short, our \ours{} distinguishes itself from existing benchmarks as follow:

\vspace{-0.15cm}

\begin{itemize}[leftmargin=*]
    \item \textbf{Cognition Inspired:} In contrast to existing benchmarks that focuses on task-level evaluation, we study the intelligence of large models from a cognition perspective. The benchmark dissects the cognitive abilities of contemporary MLLMs into five foundational factors, as per the Cattell-Horn-Carroll theory.
    This cognitive theory underpins the structure of our evaluation, informing the specific types of questions devised to test each cognitive skill. 
    \item \textbf{Multilingual Coverage:} To comprehensively measure the cognitive abilities of multimodal large models across multiple languages, \ours{} is constructed to span six languages: English, French, Chinese, Spanish, Portuguese, and Korean. In order to mitigate English-centric bias, all data relevant to cultural backgrounds have been sourced from native language resources, except for questions that transcend cultural considerations—such as the Raven test and number series problems.
\end{itemize}

The subsequent content of this section is organized as follows: In sec.~\ref{abilities}, we introduce the five-factor cognitive model of \ours{} and discuss the design philosophy behind it. In sec.~\ref{data}, we describe how we designed and collected the questions for \ours{} and provide some statistical data on \ours{}.

\begin{figure}[t]
  \centering
  \includegraphics[width=\textwidth]{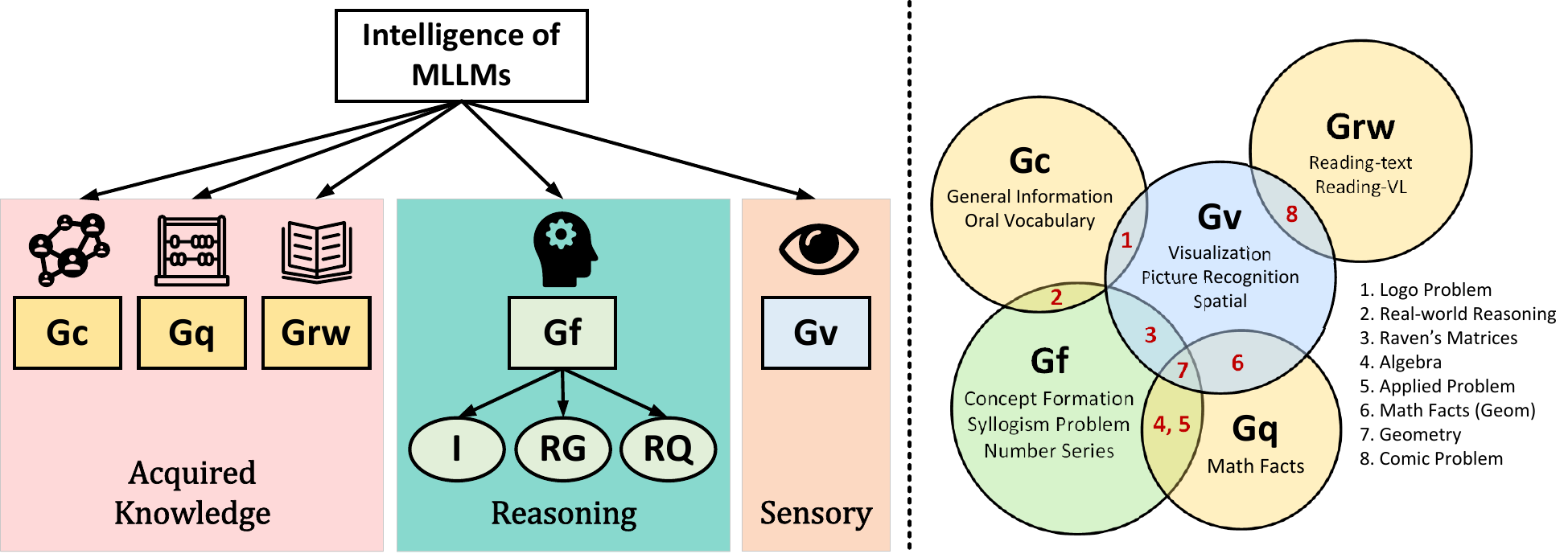}
  \caption{\textbf{Structure of our CHC inspired model of cognitive abilities.} \textbf{(Left)} We identified five key cognitive factors for current MLLMs: Comprehension-Knowledge (Gc), Quantitative knowledge (Gq), Reading and Writing (Grw), Fluid reasoning (Gf), and Visual-spatial processing (Gv). In the hierarchical structure, Gf is further subdivided into three narrow factors: I (Induction), RG (Deductive Reasoning), and RQ (Quantitative Reasoning). \textbf{(Right)} A conceptual map of the five cognitive factors and their overlaps with each other. }
  \label{chc_model}
\end{figure}

\subsection{The Five-factor Cognitive Model of \ours{}}
\label{abilities}
To formally study the large models intelligence level, we start from the state-of-the-art cognitive model, Cattell-Horn-Carroll~(CHC)~\cite{schneider2012cattell}, which is by far the most empirically validated structure model of human cognition~\cite{mcgrew2004internal}.
The CHC theory articulates a hierarchical framework of human cognitive abilities divided into three strata: general intelligence ``g'' (stratum III), broad cognitive abilities (stratum II), and narrow cognitive abilities (stratum I).
The theory has now expanded to include 16 broad abilities and over 80 narrow abilities. 
These broad but domain-specific abilities are nevertheless positively associated with one another. This positive manifold is accounted for in the CHC model by a general factor of intelligence (``g'') at stratum III.
While there is ongoing discourse regarding the exact delineation of the narrow abilities, 9 out of the 16 broad cognitive abilities have achieved substantial consensus and are well-supported by empirical evidence and practical application.\cite{caemmerer2020beyond}
These include Fluid Reasoning (Gf), Comprehension-Knowledge (Gc), Visual Processing (Gv), Auditory Processing (Ga), Short-term Memory (Gsm), Long-term Retrieval (Glr), Processing Speed (Gs), Quantitative Knowledge (Gq), and Reading and Writing Abilities (Grw).

\begin{figure}[t]
  \centering
  \includegraphics[width=\textwidth]{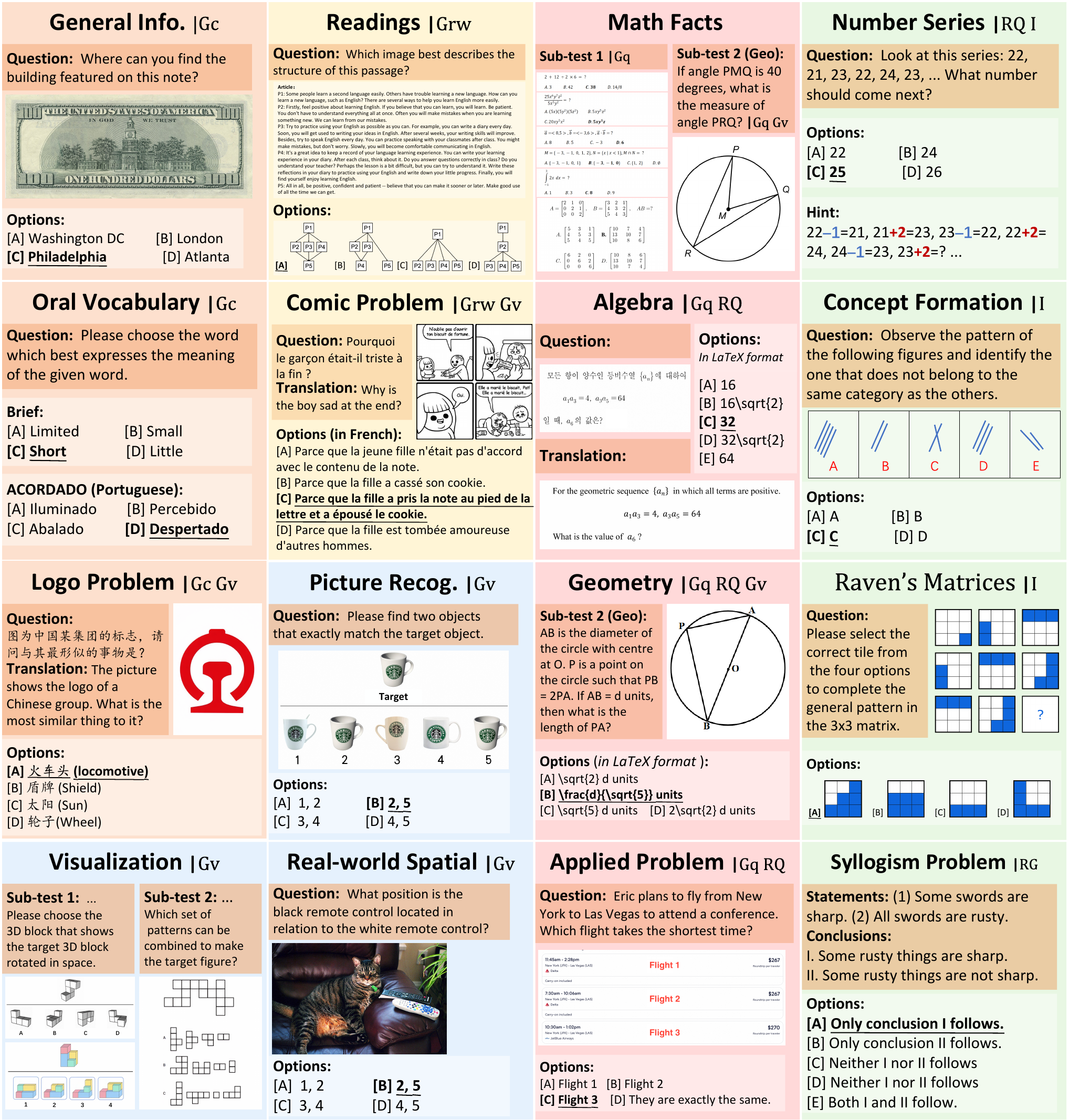}
  \caption{\textbf{Questions overview of \ours{}.} To assess the five CHC cognitive factors—Gf, Gc, Gq, Grw, Gv correspondingly—we devised five broad question clusters: common sense (orange), visual-spatial (blue), comprehension (yellow), mathematics (red), and reasoning (green). 
  To prevent the assessment of any particular ability from being constrained to a fixed and singular perspective, we have stratified each of the five clusters into 2-4 specialized narrow question types, each narrow question type reflects a different perspective on the broad CHC ability. This subdivision results in a total of 18 subtasks. 
  All the QAs are in the format of multiple choice problems whose answers are marked [A][B][C][D].}
  \label{examples}
\end{figure}

As shown in Fig.~\ref{chc_model}, the structure of our \ours{} is underpinned by the five-factor hierarchical cognitive model, which is derived from the CHC model of cognitive abilities.
Although Large Language Models exhibit cognitive processes similar to humans, they also differ in internal mechanisms, particularly with regard to processing speed (Gs) and memory (Gwm, Glr), which is greatly related to external technologies beyond the model itself, such as external databases and retrieval-augmented generation (RAG)~\cite{lewis2020retrieval}.
Additionally, given that the majority of current MLLMs, with the exception of a select few closed-source models, are not yet expanded to embrace the auditory modality, we have not included the Ga (Auditory Processing) factor in this version of \ours{}, reserving it as one of the directions for future expansion.
Consequently, based on the consultations with psychology experts, we have chosen to assess the cognitive abilities of current MLLMs in this iteration of \ours{} by focusing on five key CHC factors: Gc, Grw, Gq, Gf, and Gv, from among the nine most frequently identified CHC factors.

Interestingly, the five factors we select align closely with those of the renowned Stanford-Binet Test, Fifth Edition (SB5)~\cite{roid2012stanford}, which was also constructed upon five cognitive factors derived from the CHC theory. 
Specifically, the five cognitive factors identified in the SB5 are: Fluid Reasoning (FR), Knowledge (KN), Quantitative Reasoning (QR), Visual-Spatial Processing (VS), and Working Memory (WM). Except for Working Memory (WM), which we have substituted with Grw, these factors align directly with our selected factors, corresponding to Gf, Gc, Gq, and Gv, respectively.
This alignment is noteworthy, as the selection of these factors for the SB5 was based on extensive research on school achievement and expert ratings of the importance of these factors in the assessment of reasoning, especially in giftedness assessment~\cite{roid2004essentials}.

\subsection{Question Design and Collection}
\label{data}

Our \ours{} contains a total of 1,800 multiple choice problems, of which 1,200 are Visual Question Answering (VQA) questions. The ground truths for 630 questions are human-annotated, while the remainder of the answers for 1,170 questions were gathered from the Internet. 

As shown in Fig.~\ref{teaser}, we have devised five broad question clusters: reasoning, visual-spatial, common sense, mathematics and comprehension, separately corresponding to the assessment of the five CHC cognitive factors – Gf, Gv, Gc, Gq, and Grw. See {\em Appendix} for more details.
To prevent the assessment of any particular ability from being constrained to a fixed and singular perspective, we have stratified each of the five clusters into 2-4 narrow question types that reflect different perspectives on a broad CHC construct. This subdivision results in a total of 18 distinct question types, each designed to tap into different facets of the ability being measured. 
Consequently, any generalizations that are made from a cluster are based on two or more samples of ability, which reduces the possibility of making over-generalizations from a narrow sampling of ability. 

Moreover, as illustrated in the right part of Fig.~\ref{chc_model}, the five cognitive factors are not isolated but rather overlap with each other. For example, Fluid reasoning (Gf) not only has a process facet (inductive vs. deductive reasoning) but also has a content facet (verbal, spatial, and quantitative), each of which overlaps with other broad abilities.~\cite{schneider2018cattell}.
In order to conduct a comprehensive measurement of this overlapping nature, our narrow question types include not only tests that measure each cognitive factor individually but also cover the parts where these factors overlap. The corresponding relationships between the question types, the cognitive factors and their intersections are also shown in Fig.~\ref{chc_model}.

What is more, to ensure that our assessment remains anchored in reality, we incorporate real-world problems into the evaluations of cognitive abilities. Specifically, each broad question type includes not only abstract cognitive test questions but also typical real-world problems that require the use of one or more cognitive abilities. This approach enables us to conduct a more accurate and practical assessment of how well these abilities are applied outside controlled, test-like environments.
To ensure a balanced and comprehensive evaluation for each ability, we have tried our best to maintain an even distribution among problems associated with different abilities during data collection.

Examples of the narrow question types can be seen in Fig.~\ref{examples}, while more detailed descriptions are included in the {\em Appendix}.

\subsection{Metrics}
We use two type of metrics in our evaluation benchmark. For each narrow question type, we follow the existing benchmarks~\cite{liu2023mmbench,fu2024mme} to use accuracy. However, to holistically compare the cognitive ability, we design a novel metric general intelligence accuracy~(GIA) based on findings in cognitive field. To compute the GIA scores of the models and validate the consistency of the cognitive structure between MLLMs and human intelligence, we adopted a standard psychometric approach. This involved utilizing a confirmatory factor analysis (CFA) model, developed from our collected human evaluation data. For more details about the CFA process, see {\em Appendix} for more details.

\section{Evaluation Results}
\label{experiment}

In this section, we evaluate a total of 24 MLLMs and 480 human participants using our \ours{}. The MLLMs comprise both closed-source models, such as GPT-4o~\cite{GPT-4o}, and open-source models~\cite{liu2023improved,li2024mini,young2024yi,bai2023qwen,wang2023cogvlm,lu2024deepseek,chen2023internvl}, including LLaVA~\cite{liu2023improved} and Mini-Gemini~\cite{li2024mini}.
Our evaluation for the MLLMs is conducted under a zero-shot setting to assess the capability of models to generate accurate answers without fine-tuning or few-shot demonstrations on our benchmark. 
For all models, we conduct prompt engineering on the validation set and use the most effective prompt for the zero-shot setup in the experiments. All experiments are conducted with NVIDIA A800 GPUs~\cite{liu2023improved,li2024mini}.

\paragraph{Human Performance Baseline.} To establish a reference for human cognitive levels against MLLMs, we collected 480 valid sets of test data from human subjects using electronic questionnaires. These 480 participants were from native countries of the six selected languages, with 80 individuals per language. The 1,800 questions of \ours{} are then divided into six complete sub-questionnaires by language, with each individual only responsible for completing the sub-questionnaire corresponding to their native language. See supplementary for more details. 

\subsection{
Accuracy Score on Five Cognitive Factors
}
\label{result_chc}

\belowrulesep=0pt\aboverulesep=0pt

\begin{table}[t!]
\renewcommand{\arraystretch}{1.25}
  \centering
  \scriptsize
  \caption{\textbf{The accuracy results on 24 MLLMs regarding each cognitive ability.} The best in {\bf bold} and the second-best \underline{underlined}. All the numbers are presented in decimal and the full score is 100.
  }
  \label{tab:result1}
  \setlength{\tabcolsep}{2.45mm}{
        \begin{tabular}{>{\centering\arraybackslash}p{1.1cm}>{\raggedright\arraybackslash}p{2.85cm}>{\centering\arraybackslash}p{0.036\linewidth}>{\centering\arraybackslash}p{0.0225\linewidth}>
        {\centering\arraybackslash}p{0.0225\linewidth}>{\centering\arraybackslash}p{0.0225\linewidth}>{\centering\arraybackslash}p{0.045\linewidth}>{\centering\arraybackslash}p{0.0225\linewidth}>{\centering\arraybackslash}p{0.0225\linewidth}>{\centering\arraybackslash}p{0.0225\linewidth}>{\centering\arraybackslash}p{0.0225\linewidth}>{\centering\arraybackslash}p{0.0485\linewidth}}
        \customtop
         \multirow{2}{*}{\shortstack{Types\\(LLM Size)}}&\multirow{2}{*}{Models}&\multirow{2}{*}{\shortstack{ViT\\Size}}& \multicolumn{4}{c}{Gf} &\multirow{2}{*}{Gc}&\multirow{2}{*}{Gq}&\multirow{2}{*}{Grw}& \multirow{2}{*}{Gv}&\multirow{2}{*}{\shortstack{Overall\\Acc}}\\
         \cmidrule{4-7}&&&  I& RG&RQ& Overall& & &&\\
         
         \custommid
         \textbf{Human}        &\texttt{Average Performance}  &-      &\underline{\textbf{86.8}}    &\underline{\textbf{60.0}}    &\underline{\textbf{71.2}}    &\underline{\textbf{69.7}}    &\underline{\textbf{79.1}}    &\underline{\textbf{65.4}}    &\underline{\textbf{78.1}}    &\underline{\textbf{81.1}}    &\underline{\textbf{76.9}}\\
         
         \customhline
         \multirow{6}{*}{API}  &\texttt{GPT-4o}               &-      &\textbf{58.0}    &\textbf{59.2}    &33.9    &50.1    &72.3    &42.8    &\textbf{79.6}    &46.3    &\underline{59.8}\\
                               &\texttt{GPT-4v}               &-      &\underline{56.7}    &56.3    &\underline{40.9}    &\underline{51.9}    &\underline{74.8}    &\underline{46.4}    &\underline{77.5}    &\underline{52.4}    &59.2\\
                               &\texttt{Gemini-1.5-Pro}       &-      &54.3    &\underline{56.4}    &\textbf{41.8}    &\textbf{54.3}    &\textbf{75.8}    &\textbf{60.8}    &77.1    &\textbf{53.8}    &\textbf{62.4}\\
                               &\texttt{Gemini-Pro}           &-      &39.0    &30.8    &22.7    &32.4    &56.5    &31.7    &67.1    &43.1    &46.5\\
                               &\texttt{Cluade3-Sonnet}       &-      &39.7    &32.9    &27.3    &34.0    &58.3    &34.2    &61.3    &43.9    &47.0\\
                               &\texttt{Cluade3-Haiku}        &-      &35.3    &35.8    &30.3    &33.1    &55.8    &33.3    &57.9    &36.4    &43.1\\
         
         \customhline
                                  &\texttt{Mini-Gemini-34b}          &0.3B      &\underline{37.7}    &37.5    &30.6    &\underline{34.8}    &\underline{61.0}    &34.2    &62.9    &\underline{45.7}    &\underline{48.2}\\
         \multirow{2}{*}{OSS}     &\texttt{Mini-Gemini-8*7b}         &0.3B      &28.7    &30.0    &26.7    &30.3    &58.1    &35.0    &61.3    &41.9    &44.8\\
         \multirow{2}{*}{(Large)} &\texttt{LLaVA-v1.6-34b}           &0.3B      &20.7    &\underline{40.0}    &28.5    &30.8    &53.8    &\underline{36.4}    &61.7    &40.4    &42.8\\
                                  &\texttt{Yi-VL-34b}                &0.6B      &25.0    &32.9    &\textbf{35.8}    &29.5    &48.1    &29.2    &54.6    &35.7    &38.2\\
                                  &\texttt{InternVL-chat-v1.2-plus}  &6B        &\textbf{45.0}    &\textbf{42.5}    &\underline{32.4}    &\textbf{42.5}    &\textbf{64.6}    &\textbf{41.4}    &\textbf{66.7}    &\textbf{47.5}    &\textbf{51.9}\\
        
         \customhline
         \multirow{2}{*}{OSS}      &\texttt{Mini-Gemini-13b}        &0.3B      &\underline{22.3}    &\textbf{29.2}    &\underline{23.3}    &\textbf{24.3}    &\underline{41.5}    &\underline{26.1}    &\underline{44.2}    &28.3    &\underline{32.9}\\
         \multirow{2}{*}{(Medium)} &\texttt{LLaVA-v1.5-13b}         &0.3B      &17.7    &\underline{26.3}    &15.2    &19.9    &\textbf{42.1}    &20.3    &40.0    &\textbf{28.8}    &30.4\\
                                   &\texttt{LLaVA-v1.6-vicuna-13b}  &0.3B      &\textbf{23.3}    &19.6    &\textbf{24.5}    &\underline{23.1}    &36.7    &\textbf{26.9}    &\textbf{47.5}    &\underline{28.5}    &\textbf{33.2}\\
        
         \customhline
                                  &\texttt{Fuyu-8b}                &-         &21.7    &22.1    &27.3    &23.3    &27.3    &24.4    &27.1    &24.9    &25.1\\
                                  &\texttt{Mini-Gemini-8b}         &0.3B      &\textbf{37.3}    &\underline{29.6}    &\textbf{31.8}    &\textbf{30.4}    &\underline{51.5}    &\textbf{30.6}    &\textbf{56.3}    &\underline{36.1}    &\textbf{41.4}\\
                                  &\texttt{LLaVA-v1.5-7b}          &0.3B      &18.0    &25.0    &15.8    &19.7    &41.5    &19.7    &35.0    &25.7    &28.4\\
                                  &\texttt{LLaVA-v1.6-vicuna-7b}   &0.3B      &21.3    &22.9    &18.2    &20.5    &36.5    &19.4    &32.9    &26.9    &31.5\\
         \multirow{2}{*}{OSS}     &\texttt{LLaVA-v1.6-mistral-7b}  &0.3B      &24.3    &25.8    &24.5    &24.9    &38.5    &24.2    &36.7    &32.1    &28.9\\
         \multirow{2}{*}{(Small)} &\texttt{Deepseek-VL-7b}         &0.38B     &\underline{32.3}    &29.2    &22.1    &28.3    &50.4    &24.4    &54.2    &32.4    &37.5\\
                                  &\texttt{Yi-VL-6b}               &0.6B      &25.2    &\textbf{35.5}    &26.2    &\underline{28.8}    &35.6    &\underline{29.0}    &\underline{54.5}    &30.8    &34.4\\
                                  &\texttt{Qwen-VL}                &1.9B      &18.7    &23.8    &25.2    &22.5    &41.0    &27.5    &42.5    &30.1    &32.1\\
                                  &\texttt{CogVLM2-LLaMA3-Chinese} &10B       &29.7    &21.7    &\underline{29.7}    &26.5    &\textbf{54.8}    &27.2    &37.9    &\textbf{40.3}    &\underline{38.7}\\
         \custombot
        \end{tabular}}
\end{table}

We report the accuracy of each type of question for the 24 models alongside the average human performance for each cognitive ability in Table.~\ref{tab:result1}. We categorize the models into groups by their types, where open-source (OSS) MLLMs are grouped according to the size of their LLMs.
It's observed that even the most advanced MLLMs only marginally meet the passing line (60) for overall accuracy, e.g., Gemini-1.5-Pro (62.4) / GPT-4o (59.8) vs human (76.9).
Notably, these models excel in domains related to verbal skills and knowledge, such as Gc and Grw. This success can likely be attributed to the powerful language capabilities inherent in large language models, bolstered by their extensive training datasets.

However, a significant performance gap remains between MLLMs and humans in areas like Visual-Spatial Abilities (Gv) and Fluid Reasoning (Gf). This is particularly evident in the Visual-Spatial Abilities domain, where all models lag considerably behind human capabilities, e.g., Gemini-1.5-Pro (53.8) vs human (81.1). This underscores a substantial opportunity for advancements in the visual aspects of MLLMs. See supplementary for case studies. 
Furthermore, our findings also highlight a pronounced deficiency in the Fluid Reasoning (Gf) capability among all MLLMs, particularly in tasks involving Induction (I) and Quantitative Reasoning (RQ). 
However, it is surprising to note that in the domain of Deductive Reasoning (RG), the most advanced MLLMs, such as GPT-4o, are approaching the average human level with scores of 59.2 compared to 60.0 for human participants. This might be attributed to the strategy they use synthetic reasoning data to enhance such ability~\cite{chung2024scaling}.

Overall, MLLMs perform well in crystallized intelligence (Gc), possibly owing to their extensive training data, while the most advanced MLLMs still have a large gap with humans in fluid intelligence. This proves that our benchmark M3GIA can measure the difference between crystallized intelligence and fluid intelligence of MLLMs from a cognitive perspective, which is the key difference between M3GIA and other benchmarks.

\begin{figure}[ht]
  \centering
    \begin{minipage}[t]{0.405\textwidth}
      \centering
      \includegraphics[height=4.5cm]{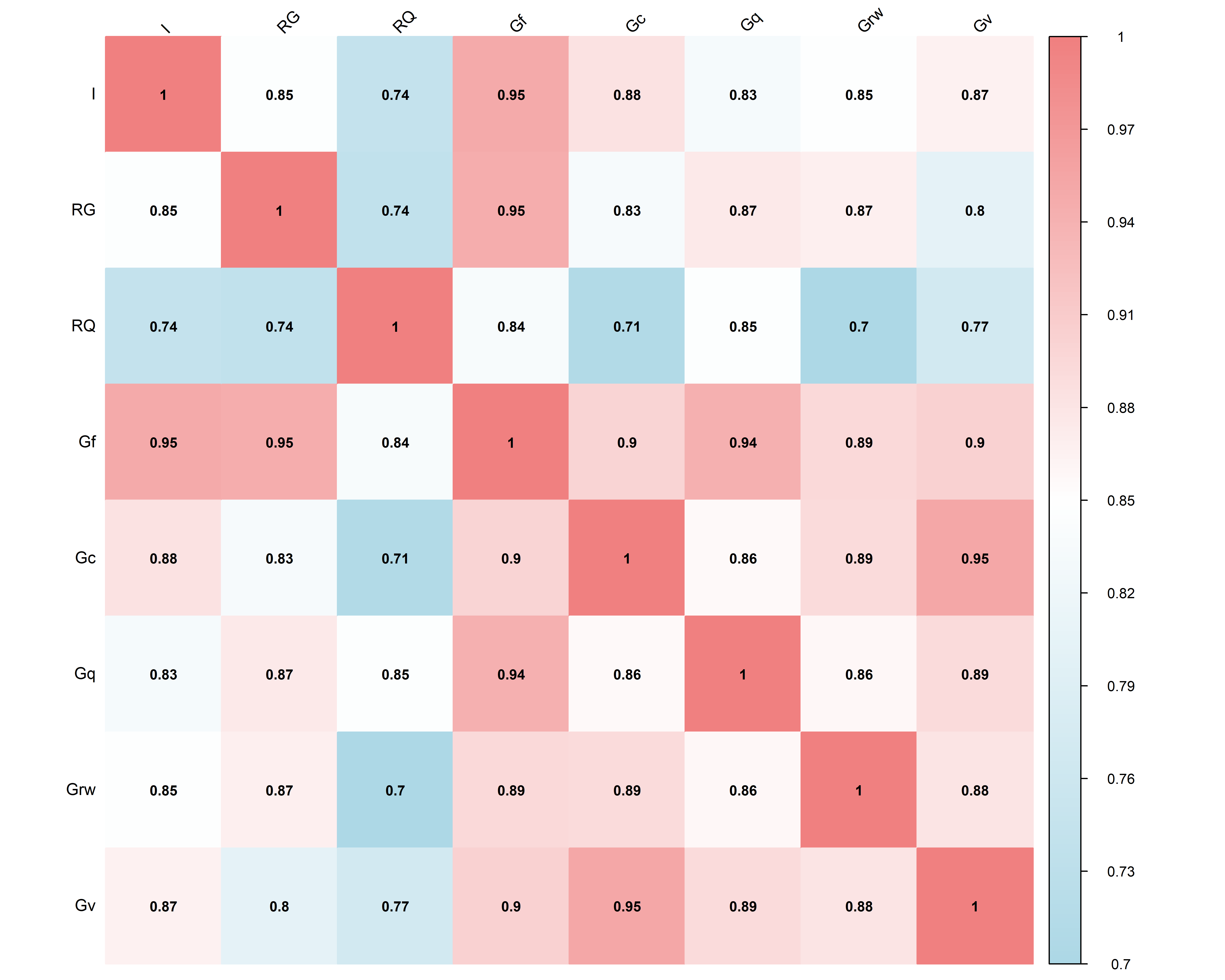}
    \end{minipage}
    \begin{minipage}[t]{0.589\textwidth}
      \centering
      \includegraphics[height=4.5cm]{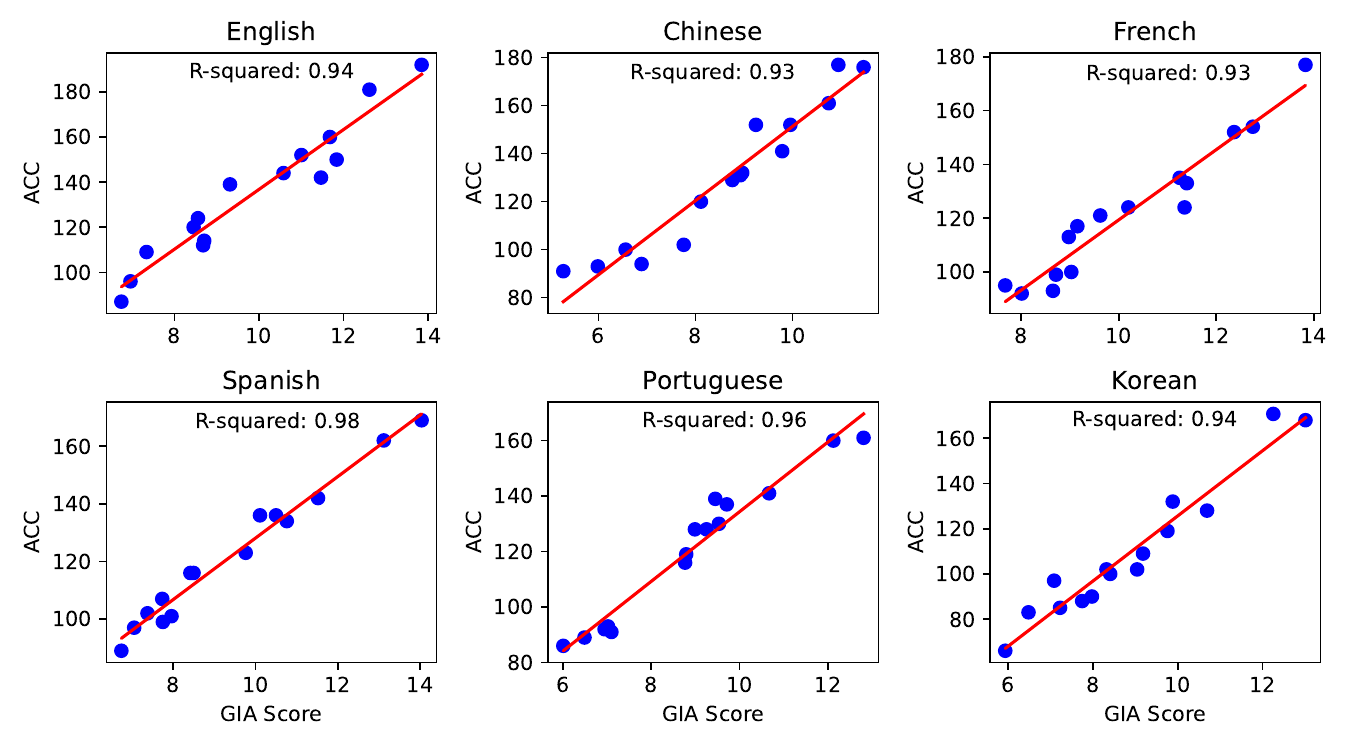}
    \end{minipage}
    
  \caption{\textbf{(Left)} The correlation matrix of MLLMs' accuracy scores across various cognitive factors. It indicates a notable correlation in their performance across these dimensions. \textbf{(Right)} We find that the GIA scores of MLLMs, calculated using the CFA model constructed from human data, demonstrate high explanatory power for their overall test performance ($R^2 \geq 0.93$).
  }
  
  \label{fig: cor}
\end{figure}

\paragraph{Winner Takes All.} More importantly, our finding reveals an intriguing \textit{Winner Takes All} phenomenon that merits further attention beyond the initial observations. Specifically, we noted a consistent trend within each group of models where proficiency in one cognitive domain often translates into superior performance across other domains as well in Table~\ref{tab:result1}. In particular, despite the diversity in score distribution among different abilities, there is a noteworthy pattern: the models achieving the top and second-best scores across various cognitive abilities are predominantly the same two models within each group.

This shows an interesting consistency to the pattern observed in human intelligence which empirically suggests the existence of General Intelligence Ability (see Sec.~\ref{related}). Therefore, it offers compelling evidence that general intelligence ability, also identified as the general factor of intelligence (``g'') at the stratum III of the CHC model, has also emerged in large models. 
Furthermore, it suggests that as MLLMs evolve towards more comprehensive cognitive processes, they too demonstrate a foundational GIA factor that simultaneously governs a variety of cognitive abilities.

\subsection{Multilingual GIA Scores}
\label{result_GIA}

\begin{figure}[ht]
  \centering
    \begin{minipage}[t]{0.394\textwidth}
      \centering
      \includegraphics[height=4.6cm]{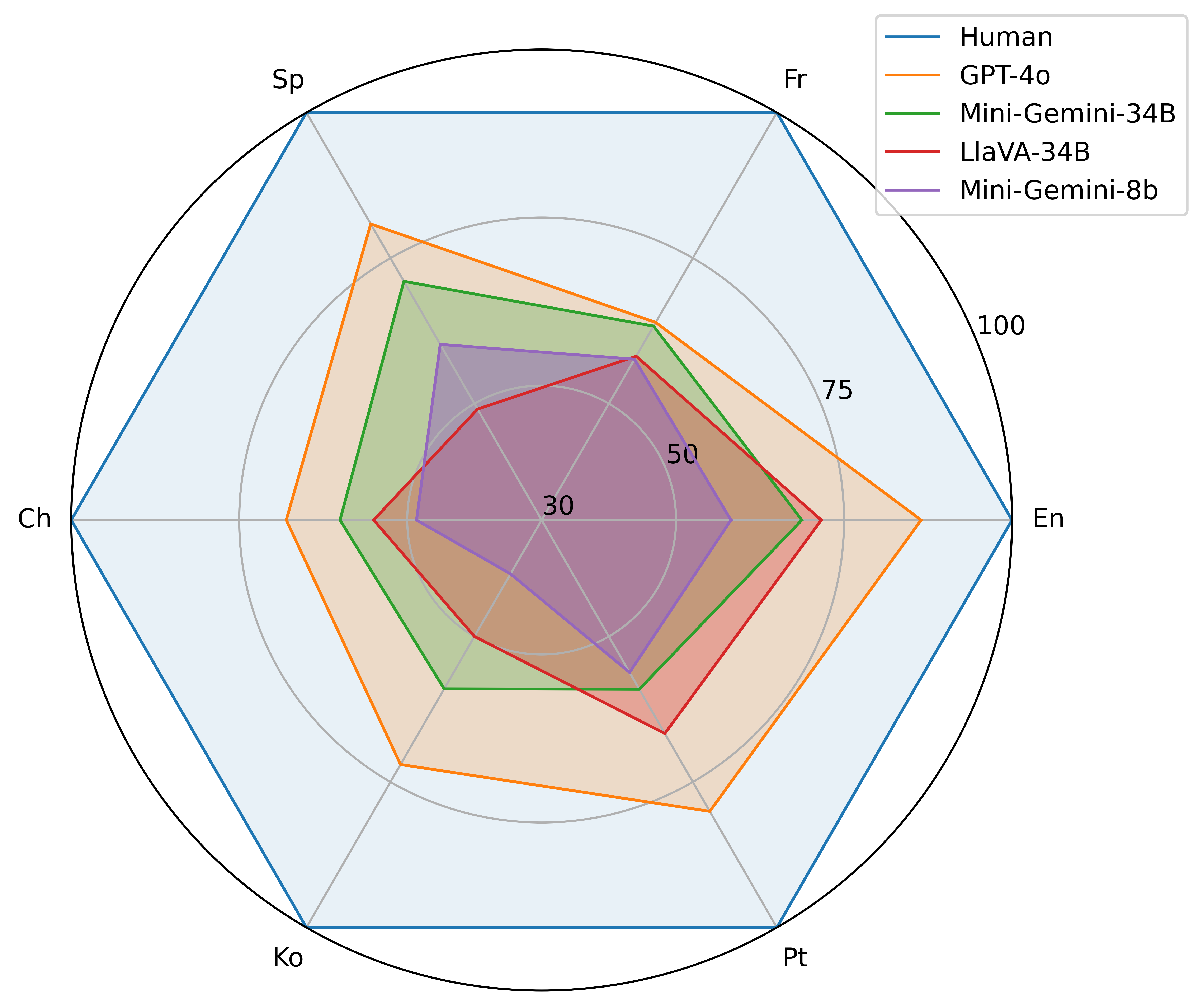}
    \end{minipage}
    \begin{minipage}[t]{0.6\textwidth}
      \centering
      \includegraphics[height=4.6cm]{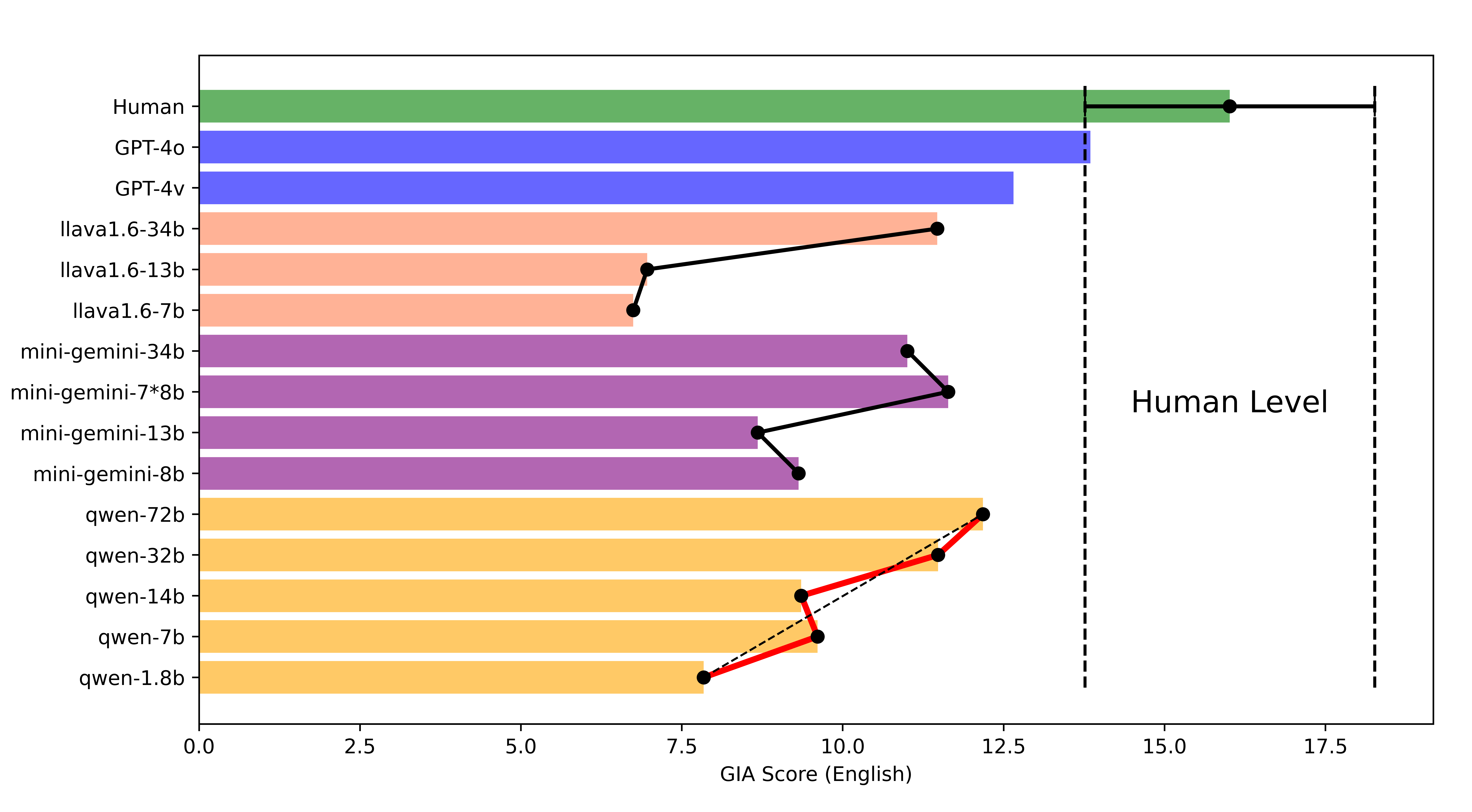}
    \end{minipage}
    
  \caption{
  \textbf{The GIA scores of some representative MLLMs across the six languages.} \textbf{(Left)} We designate the average human score as 100 in each language and normalize the scores of different models accordingly, making the GIA scores comparable across languages. \textbf{(Right)} Taking English as an example, we visualized the performance of various models and compared it with the human level.
  }
  \label{fig: GIA_en}
\end{figure}

By collecting a large amount of testing data from human subjects, we adoptted  CFA (Confirmatory Factor Analysis)  model to calculate the GIA scores which can reflect comprehensive intelligence factors. 
Since the questions for each language are not exactly the same, we need to establish a separate CFA model for each language.
To our surprise, the model built with human data showed high explanatory validity for the test results of MLLMs (cor > 0.93). This indicates, to some extent, that the cognitive structure of MLLMs indeed shows similarities to humans. 
We report the GIA scores of each language for some MLLMs of different sizes in Table.~\ref{tbl:multilingual} and Fig.~\ref{fig: GIA_en}. It's observed that the current state-of-the-art MLLMs have reached the minimum level within the human subjects' confidence interval in English. However, these MLLMs still exhibit a significant performance gap compared to humans in other languages.


\belowrulesep=0pt\aboverulesep=0pt

\begin{table}[ht]
\renewcommand{\arraystretch}{1.2}
  \centering
  \scriptsize
  \caption{\textbf{The General Intelligence Ability of different models accross the six languages.} The left side displays the actual GIA scores, while the right side shows the normalized results after setting the average human GIA scores for each language to 100.0.}
  \label{tbl:multilingual}
  \setlength{\tabcolsep}{2.4mm}{
  \resizebox{1.0\linewidth}{!}{
        \begin{tabular}{>{\centering\arraybackslash}p{1.8cm}>{\centering\arraybackslash}p{0.032\linewidth}>{\centering\arraybackslash}p{0.032\linewidth}>
        {\centering\arraybackslash}p{0.032\linewidth}>{\centering\arraybackslash}p{0.032\linewidth}>{\centering\arraybackslash}p{0.032\linewidth}>{\centering\arraybackslash}p{0.032\linewidth}|>{\centering\arraybackslash}p{0.032\linewidth}>{\centering\arraybackslash}p{0.032\linewidth}>{\centering\arraybackslash}p{0.032\linewidth}>{\centering\arraybackslash}p{0.032\linewidth}>{\centering\arraybackslash}p{0.032\linewidth}>{\centering\arraybackslash}p{0.032\linewidth}>{\centering\arraybackslash}p{0.032\linewidth}}
        
        \customtop
        \multirow{2}{*}{Models}& \multicolumn{6}{c}{General Intelligence Ability (GIA)}&\multicolumn{6}{c}{Normalized GIA Scores}\\
        \cmidrule{2-13}& En& Ch& Fr& Sp&Pt& Ko&En& Ch&  Fr&Sp& Pt& Ko\\
        
        \customhline
        Human& 16.01& 16.69& 19.52& 16.22&16.00& 18.05 
        &100.0& 100.0&  100.0&100.0& 100.0& 100.0\\
        
        \customhline
        \texttt{GPT-4o}& 13.85& 11.46& 12.37& 13.12&12.80& 13.01
        &86.5& 68.7&  63.3&80.9& 80.0& 72.1\\
        
        \texttt{GPT-4v}& 12.61& 10.95& 13.83& 14.04& 12.12& 12.25
        & 78.8& 65.6& 70.8& 86.5& 75.8&67.9\\
        
        \customhline
        \texttt{LLaVA-1.6-34b}& 11.47& 9.25& 11.35& 7.96&10.67& 9.04
        &71.6& 55.4&  58.1&49.1& 66.7& 50.1\\
        
        \texttt{LLaVA-1.6-13b}& 6.96& 6.89& 8.71& 7.75& 6.94& 7.75
        & 43.5& 41.3& 44.6& 47.8& 43.4&42.9\\
        
        \texttt{LLaVA-1.6-7b}& 6.75& 5.99& 7.67& 6.74&6.01& 5.93
        &42.1& 35.9&  39.3&41.5& 37.6& 32.9\\
        
        \customhline
        \texttt{Mini-Gemini-34b}& 11.00& 9.96& 12.75& 11.52&9.45& 10.69
        &68.7& 59.7&  65.3&71.0& 59.1& 59.2\\
        
        
        \texttt{Mini-Gemini-13b}& 8.68& 7.76& 8.65& 7.73& 7.10& 7.98
        & 54.2& 46.5& 44.3& 47.7& 44.4&44.2\\
        
        \texttt{Mini-Gemini-8b}& 9.32& 8.11& 11.25& 9.76& 8.99& 7.08
        & 58.2& 48.6& 57.6& 60.2& 56.2&39.3\\
        
        \customhline
        \texttt{Qwen-72b$^\dag$}& 11.68& 10.75& 10.20& 10.50& 9.71& 9.76
        & 72.9& 64.4& 52.2& 64.7& 60.7&54.1\\
        
        \texttt{Qwen-32b$^\dag$}& 10.58& 9.79& 9.62& 10.11& 9.25& 9.18
        & 66.1& 58.7& 49.3& 62.3& 57.8&50.9\\
        
        \texttt{Qwen-14b$^\dag$}& 8.46& 8.76& 9.15& 8.49& 8.79& 8.32
        & 52.8& 52.5& 46.9& 52.4& 54.9&46.1\\
        
        \texttt{Qwen-7b$^\dag$}& 8.56& 8.93& 8.98& 8.42& 8.77& 8.41
        & 53.4& 53.5& 46.0& 51.9& 54.8&46.6\\
        
        \texttt{Qwen-1.8b$^\dag$}& 7.34& 6.56& 8.01& 7.37& 6.49& 6.48
        & 45.8& 39.3& 41.0& 45.5& 40.6&35.9\\
        
        \custombot
        \end{tabular}}}
\end{table}

To further investigate the influence of LLM size to the GIA score, we conducted an ablation study with the Qwen series. In order to strictly control variables like different training data and ViT components. we trained the models by ourselves using the same training data for pretraining and fine-tuning and we also use the same ViT component (CLIP-ViT-L-14) in the series.
Overall, the GIA scores of the models increase with the rise in LLM parameters. However, we observed a somewhat counterintuitive phenomenon. There is often no improvement in cognitive abilities from 7B to 13B, and seems to be a emerging point of General Intelligence Ability between 13B and 34B.

\section{Conclusion}
\label{conclusion}

This paper has presented \ours{}, the first evaluation benchmark that comprehensively evaluate the cognitive abilities of MLLMs under the theoretical umbrella of the well-recognized Cattell-Horn-Carroll (CHC) Model of Intelligence.
Based on the CHC theory, we identified five key cognitive factors for current MLLMs: Fluid reasoning (Gf), Comprehension-Knowledge (Gc), Visual processing (Gv), Reading and Writing (Grw), Quantitative knowledge (Gq), and designed five broad types of questions to measure them.
In order to meet the pressing need for multilingual assessment, our evaluation data spans across six languages and are collected from native language sources, including English, Chinese, French, Spanish, Portuguese and Korean.
We conducted a series of experiments to comparative analyze the cognitive abilities of various MLLMs against human performance, and discussed how factors like the size of LLM component impact cognitive abilities.

\section{Limitations and Discussions}
\label{limitations}

This version of \ours{} does not include all the broad cognitive factors of the CHC model, such as auditory processing (Ga), olfactory processing (Go), etc.
As advancements in MLLMs incorporate a broader range of modalities, more factors from the CHC framework can be integrated, ensuring that \ours{} remains at the forefront of evaluating future generations of multimodal models.


\appendix

\section{Dataset Documentation}

\subsection{Motivation}
M3GIA is a multimodal and multilingual benchmark designed to evaluate the cognitive abilities and general intelligence of MLLMs under the theoretical underpinning of human cognition.
Instead of leveraging well-developed cognitive science to understand the intelligence of MLLMs beyond superficial achievements, existing benchmarks still mainly focus on evaluating solely on task performance. 
As described in the paper, these approaches have several limitations. We aim to bridge this gap through M3GIA, providing helpful insights into the development of artificial intelligence models with true intelligence. The creation of the dataset is funded by AI Business, Alibaba Group.

\subsection{Composition}
\begin{itemize}[leftmargin=*]
    \item M3GIA contains a total of 1,800 multiple-choice problems, of which 1,200 are Visual Question Answering (VQA), while the remaining 600 are textual questions. We ensure that all VQA tasks necessitate reliance on images for resolution and cannot be resolved with text alone (see Sec.~\ref{quality}). The ground truths for 630 questions are human-annotated, while the remainder of the answers for 1,170 questions were gathered from the Internet. M3GIA includes question sets in six languages, comprising Chinese, English, Spanish, Korean, Portuguese, and French, each with 300 questions.
    \item Each question is labeled with one or several CHC factors, with involved factors marked as `1' and non-involved factors marked as `0'. Each question is also annotated with the question cluster and the narrow question type to which it belongs, to facilitate the calculation of accuracy rates.
    \item M3GIA is self-contained. We bear all responsibility in case of violation of rights.
    \item The dataset does not contain any information that might be offensive, insulting, or threatening.
\end{itemize}

\subsection{Usage and Distribution} 
\begin{itemize}[leftmargin=*]
    \item The evaluation dataset is released at \href{https://huggingface.co/datasets/Songweii/M3GIA}{https://huggingface.co/datasets/Songweii/M3GIA}.
    \item M3GIA is released under the Apache 2.0 license.
    \item The data is saved in Parquet format, where an example is shown in the README.md file. An example code snippet is also provided showing how to read and process the data.
\end{itemize}

\subsection{Maintenance}
\begin{itemize}[leftmargin=*]
    \item M3GIA will be managed and maintained by our research group. For any questions, please contact Wei Song (\texttt{songweii@zju.edu.cn}) and Prof. Kaicheng Yu (\texttt{kyu@westlake.edu.cn}), who are responsible for maintenance.
    \item If we further expand our dataset or find any errors, we will update the dataset and results in the leaderboard accordingly. It will be updated on our website.
\end{itemize}

\section{Definitions of the CHC factors}
\label{chc}

According to the Cattell-Horn-Carroll (CHC) Model of Intelligence\cite{schneider2012cattell,schneider2018cattell}, the definitions of the five cognitive factors are as follows:

\textbf{Comprehension-Knowledge (Gc)}, also known as \emph{Crystallized Intelligence}, is the knowledge of culture that is incorporated by individuals through a process of ``acculturation''~\cite{mcgrew2009chc}. Gc is typically described as the breadth and depth of acquired knowledge of the language, information and concepts of a culture, and the application of the knowledge. Gc is primarily a store of verbal or language-based declarative (knowing what) and procedural (knowing how) knowledge acquired during general life experiences.
In short, Gc reflects the ability to apply and reason using previously learned experiences and common knowledge.~\cite{schneider2012cattell}

\textbf{Fluid Reasoning (Gf)} is the broad ability involved in reasoning, forming concepts, and solving problems using unfamiliar information or in novel situations. It includes inductive, deductive, and quantitative reasoning and \emph{is typically evident in mental operations, such as inferential reasoning, forming concepts, classification of unfamiliar stimuli and recognizing patterns.}~\cite{mcgrew2009chc,schneider2012cattell} Furthermore, there are three factors that are generally considered the hallmark indicators of Gf:

\begin{itemize}[leftmargin=*]
    \item \textbf{Induction (I).} The ability to observe a phenomenon and discover the underlying principles or rules that determine its behavior.
    \item \textbf{Deductive Reasoning (RG).} This ability, also known as general sequential reasoning, refers to the capacity to reason logically using known premises and principles step by step.
    \item \textbf{Quantitative Reasoning (RQ).} The ability to reason, either with induction or deduction, with numbers, mathematical relations, and operators.
\end{itemize}

\textbf{Visual-spatial Processing (Gv)} is the ability to perceive, analyze, synthesize, and think with visual patterns, or more succinctly, "the ability to make use of simulated mental imagery to solve problems". Once the eyes have transmitted visual information, the visual system of the brain automatically performs a large number of low-level computations (e.g., edge detection, light/dark perception, color-differentiation, motion-detection, and so forth). The results of these low-level computations are used by various higher-order processors to infer more complex aspects of the visual image.~\cite{schneider2012cattell}.
Gv abilities are typically measured by tasks (figural or geometric stimuli) that require the perception and transformation of visual shapes, forms, or images and/or tasks that require maintaining spatial orientation with regard to objects that may change or move through space.~\cite{mcgrew2009chc}

\textbf{Reading and Writing (Grw)} is the depth and breadth of knowledge and skills related to written language. It is worth noting that, although reading and writing are clearly distinct activities, the underlying sources of individual differences in reading and writing skills do not differentiate between the two activities cleanly~\cite{schneider2012cattell}. It appears that the ability that is common across all reading skills also unites all writing skills.

\textbf{Quantitative Knowledge (Gq)} is the depth and breadth of knowledge related to mathematics. Specifically, it is the ability to comprehend quantitative concepts and relationships and to manipulate numerical symbols. It consists of acquired knowledge about mathematics such as knowledge of mathematical symbols (e.g., $\int, \pi, \sum, \infty, \neq, \leq, +, -, \times, \div,$ and many others), operations (e.g., addition/subtraction, multiplication/division, exponentiation/nth rooting, factorials, negation, and many others), computational procedures (e.g., long division, reducing fractions, quadratic formula, and many others).
Gq abilities are typically measured by tests include measures of math calculation, applied problems (or math problem solving), and general math knowledge (e.g., Arithmetic on the Wechsler Scales, Quantitative Reasoning on the SB5).

\section{Introduction to the Evaluation Questions}
\label{question}

\belowrulesep=0pt\aboverulesep=0pt

\begin{table}[t!]
\renewcommand{\arraystretch}{1.5}
  \centering
  \footnotesize
  \caption{\textbf{The number and cognitive factors of each question type.} Our M3GIA is organized into five clusters, each cluster is further defined to combine two or more narrow question types that are aspects of a broad CHC construct (real-world problems in {\bf bold}). In total, it contains a total of 1,800 meticulously designed multilingual questions, with the number of questions and the distribution of question types being completely consistent across different languages. 
  Questions potentially related to the cultural backgrounds are marked in \textcolor{gg}{green}, while purely intellectual questions, unrelated to cultural background, are marked in \textcolor{rr}{yellow}. The former's data are sourced from native language context, while the latter uses questions translated into the six languages.}
  \label{tab:design}
  \setlength{\tabcolsep}{1.2mm}{
        \begin{tabular}{>{\centering\arraybackslash}p{20.4mm}>{\centering\arraybackslash}p{30mm}|>{\centering\arraybackslash}p{0.054\linewidth}|>{\centering\arraybackslash}p{0.054\linewidth}|>{\centering\arraybackslash}p{0.054\linewidth}|>{\centering\arraybackslash}p{0.054\linewidth}|>{\centering\arraybackslash}p{0.054\linewidth}|>{\centering\arraybackslash}p{0.054\linewidth}|>{\centering\arraybackslash}p{0.054\linewidth}|>{\centering\arraybackslash}p{0.068\linewidth}}
        \toprule
        \multirow{2}*{Cluster}&\multirow{2}*{Question Types}&\multirow{2}*{Gc}&\multirow{2}*{Gv}&\multirow{2}*{Grw}&\multirow{2}*{Gq}&   \multicolumn{3}{c|}{Gf}& \multirow{2}*{Num}\\
        \cmidrule{7-9}&&& & & &    I&RG&RQ\\
        
        \hline
        \multirow{3}*{Common Sense}
        &General Information&\cellcolor{bb!30} \checkmark&&&& & & &\cellcolor{gg!30}$20\times6$\\
        &Oral Vocabulary&\cellcolor{bb!30} \checkmark   &&&&&&&\cellcolor{gg!30}$15\times6$\\
        &\textbf{Logo Problem}&\cellcolor{bb!30} \checkmark   &\checkmark&&&&&&\cellcolor{gg!30}$15\times6$\\
         
        \hline
        \multirow{3}*{Visual-spatial}
        &Visualization&&\cellcolor{bb!30}\checkmark&& &&&&\cellcolor{rr!30}$30\times6$\\
        &Picture Recognition&&\cellcolor{bb!30} \checkmark&&&&&&\cellcolor{rr!30}$15\times6$\\
        &\textbf{Real-world Spatial}&&\cellcolor{bb!30} \checkmark&&&&&&\cellcolor{gg!30}$15\times6$\\
         
        \hline
        \multirow{3}*{Comprehension}
        &Readings-text&&&\cellcolor{bb!30} \checkmark&&&&&\cellcolor{gg!30}$15\times6$\\
        &Readings-VL&&&\cellcolor{bb!30} \checkmark&&&&&\cellcolor{gg!30}$10\times6$\\
        &\textbf{Comic Problem}&&\checkmark&\cellcolor{bb!30} \checkmark&&&& &\cellcolor{gg!30}$15\times6$\\
         
        \hline
        \multirow{4}*{Mathematics}
        &Math Facts&&&&\cellcolor{bb!30} \checkmark&&&&\cellcolor{gg!30}$25\times6$\\ 
        &Algebra&&&&\cellcolor{bb!30} \checkmark&&&\checkmark&\cellcolor{gg!30}$15\times6$\\
        &Geometry&&\checkmark&&\cellcolor{bb!30} \checkmark &&&\checkmark&\cellcolor{gg!30}$10\times6$\\
        &\textbf{Applied Problem}&\checkmark&&&\cellcolor{bb!30} \checkmark &&&\checkmark&\cellcolor{gg!30}$10\times6$\\
        
        \hline
        \multirow{5}*{Reasoning}
        &Number Series&&&&&\cellcolor{bb!30} \checkmark&&\checkmark&\cellcolor{rr!30}$20\times6$\\
        &Concept Formation&&&&&\cellcolor{bb!30} \checkmark&&&\cellcolor{rr!30}$20\times6$\\
        &Raven's Matrices&&\checkmark&&&\cellcolor{bb!30} \checkmark&&&\cellcolor{rr!30}$10\times6$\\
        \cmidrule{2-10}
        &Syllogism Problem&&&&&&\cellcolor{bb!30} \checkmark&&\cellcolor{gg!30}$20\times6$\\
        &\textbf{Real-world Reasoning}&\checkmark&&&&&\cellcolor{bb!30} \checkmark&&\cellcolor{gg!30}$20\times6$\\
        
        \bottomrule
        \end{tabular}}
\vspace{-0.2cm}
\end{table}

In this section, we will outline the five question clusters and the 18 narrow question types they encompass.

\vspace{-0.2cm}

\paragraph{The Common Sense Cluster.}
The common sense cluster is designed to measures the Gc factor of an MLLM and includes 3 narrow question types: general information, oral vocabulary and logo problem. 
In \textbf{general information}, the model is presented with an image and is asked, “Where would you find [the object] in the picture?” or “What would you do with [the object] in the picture?” The initial items in each subtest draw from familiar everyday objects, and the items become increasingly difficult as the objects become more obscure or less familiar.
\textbf{Oral vocabulary} consists of two subtests: Synonyms and Antonyms. In the Synonyms subtest, the model is provided with a word and is asked to choose its synonym. In the Antonyms subtest, the model is provided with a word and is asked to choose its antonym. In CHC theory, this test primarily measures a narrow aspect of Comprehension-Knowledge (Gc) referred to as lexical knowledge (VL; vocabulary knowledge), or knowledge of words and word meanings. \cite{schrank2016essentials}
The \textbf{logo problem} is the real-world problem of the cluster, where a model is provided with a logo and is required to identify an abstract element within it. To achieve this, it must have a very deep impression on the element, such as a confusing artistic characters or symbolic expression of cultural elements, which requires a high level of Gc and a certain level of Gv.

\vspace{-0.2cm}

\paragraph{The Visual-spatial Cluster.}
This cluster is designed to evaluate the Gv factor and includes 3 narrow question types: visualization, picture recognition and real-world spatial.
\textbf{Visualization} consists of two subtests: Block Rotation and Spatial Relations. In the former, the model is asked to identify the rotated 3D block that match the original 3D block. In the latter, the model is required to identify three or four pieces that form a complete target shape.
In \textbf{picture recognition}, a model is asked to identify a subset of specified pictures within a field of distracting pictures. The stimuli and distracters for each item include varieties of the same type of object (e.g., several different leaves) to eliminate verbal mediation as a memory strategy~\cite{schrank2018woodcock}.
\textbf{Real-world spatial problem} necessitates that the model accurately determines the relative 3D positioning of objects within an image depicting real-world scenarios. This requires the model to recognize and interpret all existing relationships in the physical world, including comprehensive 3D spatial relationships and the dynamic interconnections between the objects portrayed.

\vspace{-0.2cm}

\paragraph{The Comprehension Cluster.}
This cluster is designed to evaluate the Grw factor and includes 3 narrow question types: readings-text, readings-VL and the comic problem. 
In \textbf{readings-text}, the model is provided with long articles (about 4-6 paragraphs) and will be required to answer questions related to the main ideas of the articles or the relationships between paragraphs. The articles are collected from reading comprehension exercises found in middle and high school levels across the six countries. 
To highlight the multimodal nature of our benchmark, we designed \textbf{readings-VL}, where responses must be selected from image-based options besides the conventional text-based queries.
In the \textbf{comic problem}, the model will be provided with a comic consisting of four or more panels that make up a complete plot. To answer the questions, the model needs to understand the entire story's connotation based on the textual dialogues between characters and the plot development. This approach evaluates the model's ability to integrate visual narrative comprehension with textual comprehension, challenging it to understand scenarios represented both visually and textually.

\vspace{-0.2cm}

\paragraph{The Mathematics Cluster.}
This cluster is designed to evaluate the Gq factor and includes 4 narrow question types.
\textbf{Math facts} is tailored to measure Gq alone and consists of two subtests: symbolic knowledge and geometric knowledge. The former focuses on the model's acquired knowledge about mathematical symbols and operations. It covers knowledge from elementary to university level, including arithmetic, vector operations, calculus, etc. The latter emphasizes the model's capability to solve problems using geometric knowledge. 
In \textbf{algebra} and \textbf{geometry}, we source the questions from authentic middle school and high school exam papers across the six countries. Unlike math facts problem which can be directly answered once the knowledge is acquired, these problems require a further reasoning process. Thus, they not only call upon Gq but also require RQ. 
To evaluate the model's ability to solve mathematical problems in real-life scenarios, we have specially designed \textbf{application problems}. For example, the model might be provided with a restaurant bill and asked to calculate the total amount to be paid. 
Since it rely heavily on common knowledge, Gc is also annotated in this type of problems.

\vspace{-0.2cm}

\paragraph{The Reasoning Cluster.}
This cluster is designed to assess the Gf factor and includes five narrow question types. Specifically, \textbf{number series}, \textbf{concept formation}, and \textbf{Raven's Matrices} are targeted at evaluating the I (inductive) factor, while the \textbf{syllogism problem} and \textbf{real-world reasoning} target the RG (deductive reasoning) factor.
In \textbf{number series}, the model is presented a series of numbers with one or more numbers missing. The model must determine the numerical pattern and provide the missing number in the series.
\textbf{Concept formation} measures the ability to categorize and compare~\cite{andrewes2015neuropsychology}, a basis for abstracting concepts~\cite{wang2019concept}. It requires the model to examine a series of shapes or pictures and then formulate a rule that applies to the item and then figure out the item that do not coincide with the rule.
The \textbf{syllogism problem} is a classic form of deductive reasoning, where the model is presented with two statements followed by two conclusions. The model have to take the statements to be true even if they appear to contradict commonly known facts. Then it is asked to decide which of the given conclusions logically follows from the two given statements, disregarding commonly known facts. \textbf{Real-world reasoning} refers to logical reasoning questions rooted in real-world scenarios, where Gc is also important.

\section{Data Curation Process}
\label{data_curation}

\subsection{Data Collection and Statistics}
\label{statistics}

\begin{figure}[t]
  \centering
  
  \includegraphics[width=\textwidth]{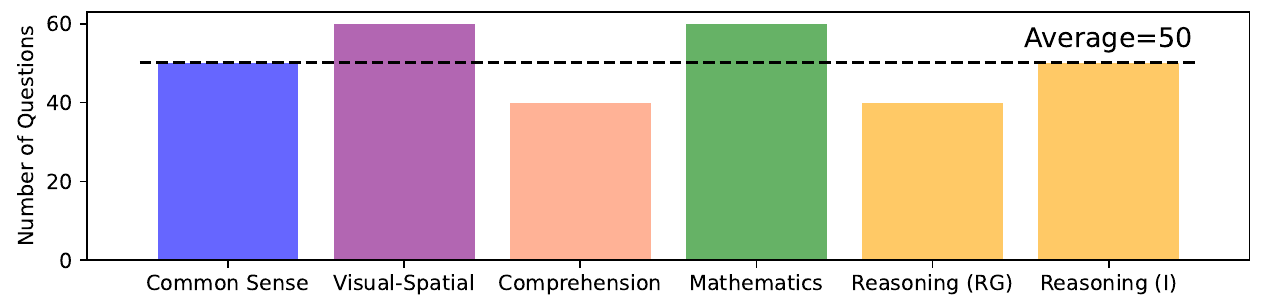}
  \caption{\textbf{Data Balancing.} we keep the number of questions for each cluster as balanced as possible when collecting questions. Given the unique characteristics of Gf, we have divided the Reasoning Cluster into Reasoning (I) and Reasoning (RG) for statistical analysis. Across each language, the number of questions within each cluster varies from 40 to 60, with an average of 50.}
  
  \label{balance}
\end{figure}


    
    
    

\paragraph{Data Balancing.}
    To ensure equal consideration for each CHC factor during the assessment, we have maintained a balanced number of questions for each cluster that measures the various CHC factors, as shown in Fig.~\ref{balance}. Specifically, the number of questions in each cluster fluctuates around 50, with a maximum capped at 60 and a minimum threshold of 40.

\vspace{-0.25cm}
\paragraph{Questions Crafted from Scratch.}
    Due to the fact that many human intelligence tests are not open to the public, and considering the novelty of some of our question types (such as logo problem, comic problem, etc.), we could not source pre-existing QA pairs from available datasets for many questions. Consequently, we have crafted numerous questions from scratch.
    For these questions, ensuring the correctness of the answers and the clarity of the descriptions is particularly important. See later Sec.~\ref{quality} for more detailed information.

\vspace{-0.25cm}
\paragraph{English-centric Bias.}
    Apart from questions that are completely independent of cultural background, such as Number Series and Raven’s Matrices, all data are sourced from native websites corresponding to the language.
    These data encompass not only text explicitly linked to cultural backgrounds but also images, since images can also convey information about the cultural contexts implicitly, such as the attire of people in the image background, architectural styles specific to a region, etc.

\vspace{-0.25cm}
\paragraph{Multimodal Nature.}
    As a multimodal benchmark, safeguarding the dataset's multimodal attributes is crucial. In particular, questions related to images should require the visual information for resolution and not be solvable through text alone. This principle was rigorously adhered to during the data collection phase, and we also placed emphasis on it during the checking process (see later Sec.~\ref{quality}). We further validated the importance of image information in our benchmark through an experiment that involved removing images from the evaluation dataset, as shown in Fig.~\ref{no_img}.


\begin{table}[h]
    \centering
    \small
    
    \caption{\textbf{Comparison of GPT-4v's accuracy rates} across five clusters \textbf{before and after the exclusion of images} from the evaluation dataset. Removing images from the dataset resulted in a notable decline in evaluation performance, underscoring the significance of visual information in the assessment and emphasizing the multimodal nature of our M3GIA benchmark.}
    \label{no_img}

    \vspace{-0.4cm}

    \begin{center}
        \renewcommand{\arraystretch}{1.2}
        \setlength{\tabcolsep}{1.5mm}{
        \resizebox{\linewidth}{!}{
        \begin{tabular}{c|cccccc}
            \customtop
                           &Common Sense &Visual-Spatial &Comprehension &Mathematics &Reasoning &Overall \\
            \custommid
            With Images    &87.0 \%      &48.0 \%        &77.8 \%       &46.4 \%     &56.5 \%   &60.7 \%\\ 
            \customhline

            Without Images &44.5 \% ${(\downarrow)}$      &23.9 \% ${(\downarrow)}$        &50.4 \% ${(\downarrow)}$       &31.7 \% ${(\downarrow)}$     &37.8 \% ${(\downarrow)}$   &36.6 \% ${(\downarrow)}$ \\
            \custombot
            
        \end{tabular}}}
    \end{center}
\end{table}


\subsection{Data Quality Control}
\label{quality}

\begin{figure}[t]
  \centering
  
  \includegraphics[width=\textwidth]{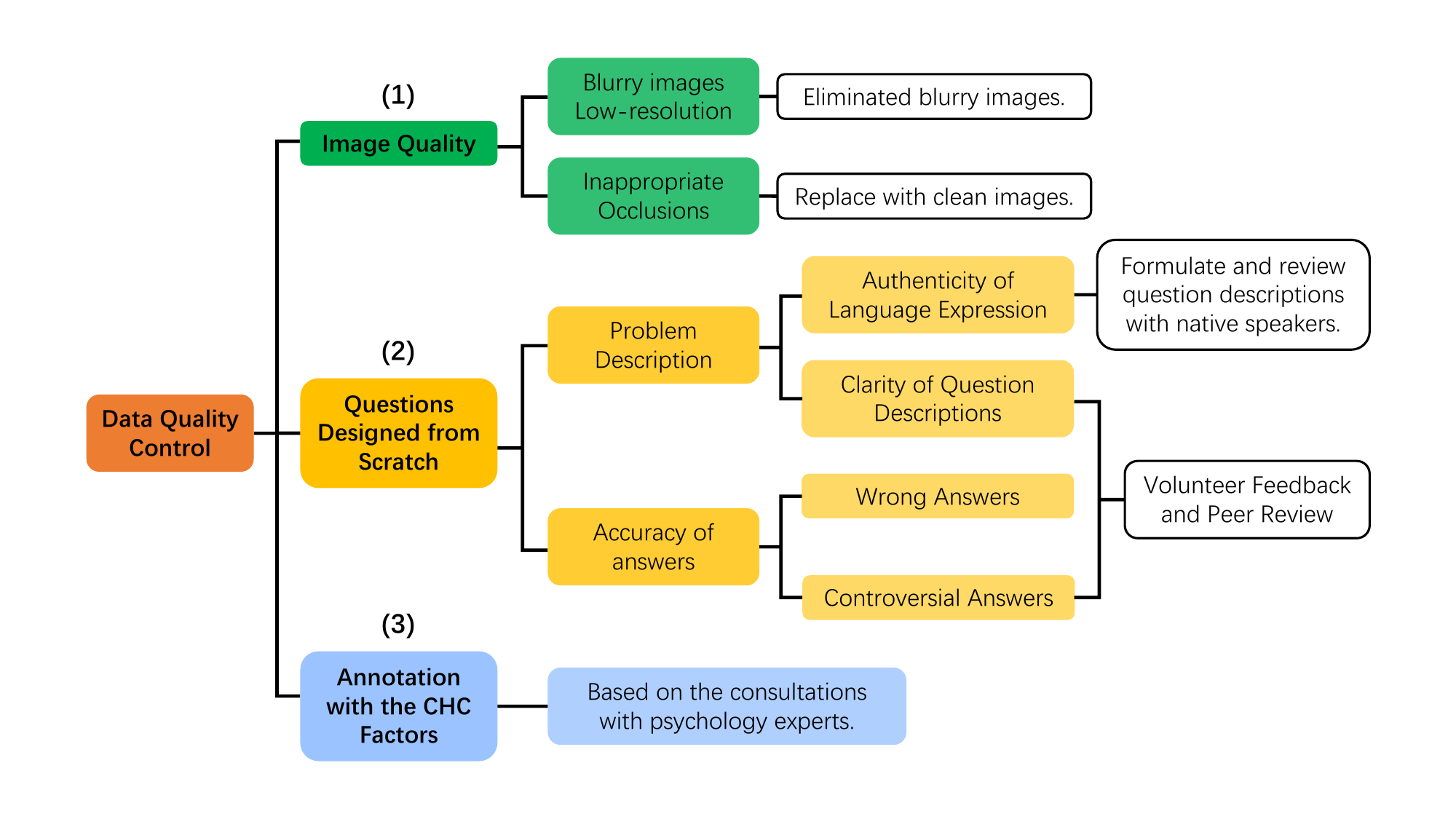}
  \caption{\textbf{How we ensure the data quality of M3GIA.}}
  
  \label{data_quality}
\end{figure}

To further control the quality of our data, we perform the data cleaning process from three perspectives, as illustrated in Fig.~\ref{data_quality}. 

\begin{itemize}[leftmargin=*]

    \item \textbf{Image Quality.}
    We traverse the dataset and locate all blurry images with resolutions lower than 100$\times$100 px.  For questions featuring these images, we either replace them with similar questions that use high-resolution images or substitute the images with clear alternatives that convey the same meaning.
    
    \item \textbf{Accuracy Check.} 
    For the questions we designed from scratch, we have paid special attention to ensuring their correctness.
    
    (i) To guarantee the authenticity of the language expression in our questions, we engaged native speakers to both formulate and review the descriptions of the question stems. Specifically, after establishing the intended meaning and creating a draft version, these native speakers undertake a thorough review, culminating in the finalized version of the question descriptions.
    
    (ii) We employed volunteer feedback and peer review as methods to assess the clarity of our question descriptions and to detect any potential issues with the answers.
    
    \emph{Clarity of Descriptions:} We recruited 10 volunteers for each language who were not involved in question creation to take our tests and provide feedback on any errors or unclear descriptions they encountered in the questions. After thorough discussion of their feedback, we ultimately incorporated revisions into 28 questions.
    
    \emph{Correctness of Answers:} After the volunteers submit their answers to the electronic questionnaire, the correct answers will be automatically disclosed. They will then be prompted to revisit any questions they answered incorrectly and are encouraged to challenge these, offering feedback on any they assert to be correct or view as contentious. This feedback was taken seriously, and we ultimately made corrections to six instances where we recognized that the answers were indeed controversial or misleading.
    Besides, we also employed peer review within our group to ensure the correctness of answers. Specifically, after formulating their questions, team members will swap them with each other for a round of testing. Following this exercise, if a tester has a justifiable reason for an incorrect response, they will engage in a direct discussion with the question's author. This method led to the identification of around ten answers that were deemed contentious.
    
    \item \textbf{Annotation of the CHC Factors.} To ensure the rationality of the questions designed for each CHC factor and the validity of the CHC factors annotated for each question, psychologists were deeply involved and cooperated in the question design and annotation phases.
    
\end{itemize}


\section{The GIA Metrics}
\label{gia}

\subsection{Human Data Collection}
\label{human}

We collected human data in each language from 80 participants using paid electronic questionnaires. 
Participation in the test is compensated and entirely voluntary. To protect user privacy, the test is also conducted anonymously.
Each participant was mandated to answer all questions to be eligible for payment. To motivate participants to provide thoughtful responses, the compensation is structured incrementally, increasing with the number of questions answered accurately. Additionally, to mitigate the risk of participants choosing answers at random just for the monetary incentive, we randomly inserted several ``check question'' within the questionnaire. For instance, a check question might instruct participants to ``Please select option B.''  If a participant answer more than two such questions incorrectly, their submission would be considered invalid.

\subsection{Calculation of the GIA Score}
\label{calculation}

In this study, we employed a cognitive factor analysis (CFA) approach to model the General Intelligence Ability (GIA) of human subjects based on the CHC theory of cognitive abilities~\cite{dubois2018distributed}. The CHC theory posits a hierarchical structure of cognitive abilities, encompassing broad factors such as Gc, Gf, Gv, Gq, and Grw, which are further broken down into narrower tasks. Our MarcoBench, a comprehensive set of 1,800 multiple choice problems corresponding to the assessment of the five CHC cognitive factors, was meticulously subdivided into 18 distinct question types, each designed to measure different facets of the cognitive abilities being assessed.

Data collection involved 80 human subjects across four different languages: Chinese, English, Portuguese, and Korean. A total of 60 subjects were utilized for model building, while the remaining 20 subjects were reserved for model validation. Subjects were administered the MarcoBench, and their performance on the tasks was meticulously recorded. The data comprised accuracy scores on 18 cognitive tasks, representing the 18 distinct question types. The accuracy data was firstly normalized to generate z-scores. And then, the EFAtools package was employed to scale the data and calculate the correlations between the variables. A series of statistical tests, including Bartlett's test and the Kaiser-Meyer-Olkin (KMO) measure, were conducted to assess the suitability of the data for factor analysis. An overall KMO value larger than 0.6 was deemed acceptable for factor analysis~\cite{watkins2018exploratory}.

The CFA model was constructed in accordance with the CHC theory, with the broad and narrow factors defined as per the theoretical framework. We used the lavaan package (https://www.lavaan.ugent.be/) to fit the CFA model to the pre-processed data. The CFA model structure included:

\begin{itemize}
    \item Gc: Measured through general information, oral vocabulary, and logo problem tasks.
    \item Gv: Included visualization, picture recognition, and real-world spatial tasks.
    \item Grw: Assessed through readings-text, readings-visual-language (VL), and comic problem.
    \item Gq: Comprised math facts, algebra, geometry, and application problems.
    \item Gf: Evaluated through number series, concept formation, Raven’s Matrices, syllogism problem, and real-world reasoning tasks.
\end{itemize}

Additionally, a General Intelligence Ability (GIA) factor was included, integrating all five broad factors. Model estimation was performed using Maximum Likelihood with Restricted Maximum Likelihood (MLR) estimation, which has been demonstrated to be more robust in the presence of multicollinearity.

The model's fit was evaluated using a range of indices, including the chi-square statistic, degrees of freedom, p-value, Comparative Fit Index (CFI), Root Mean Square Error of Approximation (RMSEA), Standardized Root Mean Square Residual (SRMR), and Akaike Information Criterion (AIC). The primary focus was on the CFI and SRMR, as they are considered more reliable indicators of model fit. A CFI larger than 0.8 or 0.9 was considered acceptable, while an SRMR equal to or lower than 0.08 was deemed acceptable~\cite{baumgartner1996applications, doll1994confirmatory}.

Upon establishing a satisfactory model fit, we employed it to calculate latent scores for the GIA on a separate set of test data. Subsequently, we calculated the Pearson correlation coefficient between the GIA latent score and the overall accuracy of the subjects on the test data to validate the model's effectiveness. The results of this analysis provided robust evidence for the validity of the CFA model in capturing the GIA of human subjects, as indicated by the significant positive correlation between the GIA latent score and overall accuracy. This validation process underscores the model's theoretical grounding in the CHC theory and its empirical support from the data. Subsequently, we applied the CFA model to estimate the GIA for several MLLMs, including gpt-4o~\cite{GPT-4o}, gpt-4v~\cite{achiam2023gpt}, llava1.6-34b~\cite{liu2023improved}, llava1.6-13b, llava1.6-7b, mini-gemini-34b~\cite{li2024mini}, mini-gemini-7*8b, mini-gemini-13b, and mini-gemini-8b, enabling a comparative analysis of their cognitive abilities against human performance.

\section{Evaluation Strategy}
\label{evaluation}

\paragraph{Option Extraction} For choice extraction, we adopted a two-stage strategy. In the first stage, we employed a keyword-based rule method to parse the model output in order to obtain options. This approach proved very effective, with the majority of existing multimodal large models successfully identifying correct answers at this stage. Yet, to enhance the robustness of our evaluation, we adopted a second stage of precautionary measures in case the parsing in the first stage fails. 
This involves deploying GPT-4-turbo for the concise summary of answer choices from the original model responses. If the second stage still fails, we will randomly generate an option for the model as the answer to the question. 
It is noteworthy, though, that throughout the actual testing process thus far, we have not encountered scenarios necessitating the use of random option generation.

The rationale behind not directly resorting to large language models for option extraction in the first stage stems from the superior stability and reliability of the rule-based method. Despite leveraging large language models for option extraction has been a common practice model evaluations, it still carries a certain error rate. On the contrary, the rule-based method, while not infallible in parsing answers across all scenarios, nearly guarantees correctness in the instances where parsing is successful. Consequently, we advocate for an initial screening using the rule-based method, followed by the employment of large language models for extraction, as a strategy that enhances overall robustness.

\paragraph{Scoring} In addition to the calculation of the GIA score mentioned above, our benchmark can also be broken down to calculate accuracy across various cognitive dimensions. Specifically, each question is annotated with the CHC factors it involves; factors that are involved are marked with a 1, and those that are not involved are marked with a 0. When a question involves a certain factor, the correctness of that question will contribute to the accuracy statistics for that particular CHC factor; otherwise, it will not be included in the statistics. Taking the calculation of the accuracy score of the Gc factor as an example:

\begin{equation}
    \begin{aligned}
        Acc\_Gc = \frac{\sum_{i=1}^{n} {Gc}_i \cdot T_i}{\sum_{i=1}^{n} Gc_i}
    \label{eq:acc_score}
    \end{aligned}
\end{equation}

where $n$ is the total number of questions, $Gc_i$ indicates whether the $i^{th}$ question involves the Gc factor, marked as 1 if it does, and 0 otherwise. $T_i$ indicates whether the $i^{th}$ question was answered correctly, with 1 representing a correct answer and 0 representing an incorrect answer.
To mitigate the effects of randomness on the evaluation results, including both the scores of the various CHC factors and the overall GIA score, we adopt a strategy of iterating five times and taking the average.

\section{GIA Scores on More Languages}
\label{experiments}

\subsection{Ablation Study on LLM Size}
\label{size}

\belowrulesep=0pt\aboverulesep=0pt

\begin{table}[ht]
\renewcommand{\arraystretch}{1.2}
  \centering
  \scriptsize
  \caption{\textbf{The training data and hyperparameters of MLLM with Qwen series.}}
  \label{tab:hyperparameters}
  \setlength{\tabcolsep}{2.4mm}{
        \begin{tabular}{>{\centering\arraybackslash}p{3.0cm}|>{\centering\arraybackslash}p{1.8cm}>{\centering\arraybackslash}p{1.8cm}}
        
        \customtop
        \multirow{1}{*}{Data and Hyperparameters}&\multirow{1}{*}{Pretrain}&\multirow{1}{*}{Finetune}\\
        
        \customhline
        data size& 558K& 1550K\\
        
        \texttt{batch size}& 256& 128\\
        
        \texttt{lr}& 1e-3& 2e-5\\
        
        \texttt{lr schedule}& cosine decay& cosine decay\\
        
        \texttt{lr warmup ratio}& 0.03& 0.03\\
        
        \texttt{weight decay}& 0& 0\\
        
        \texttt{epoch}& 1& 1\\
        
        \texttt{optimizer}& AdamW& AdamW\\
        
        \custombot
        \end{tabular}}
\end{table}

\begin{figure}[t]
  \centering
  \includegraphics[width=\textwidth]{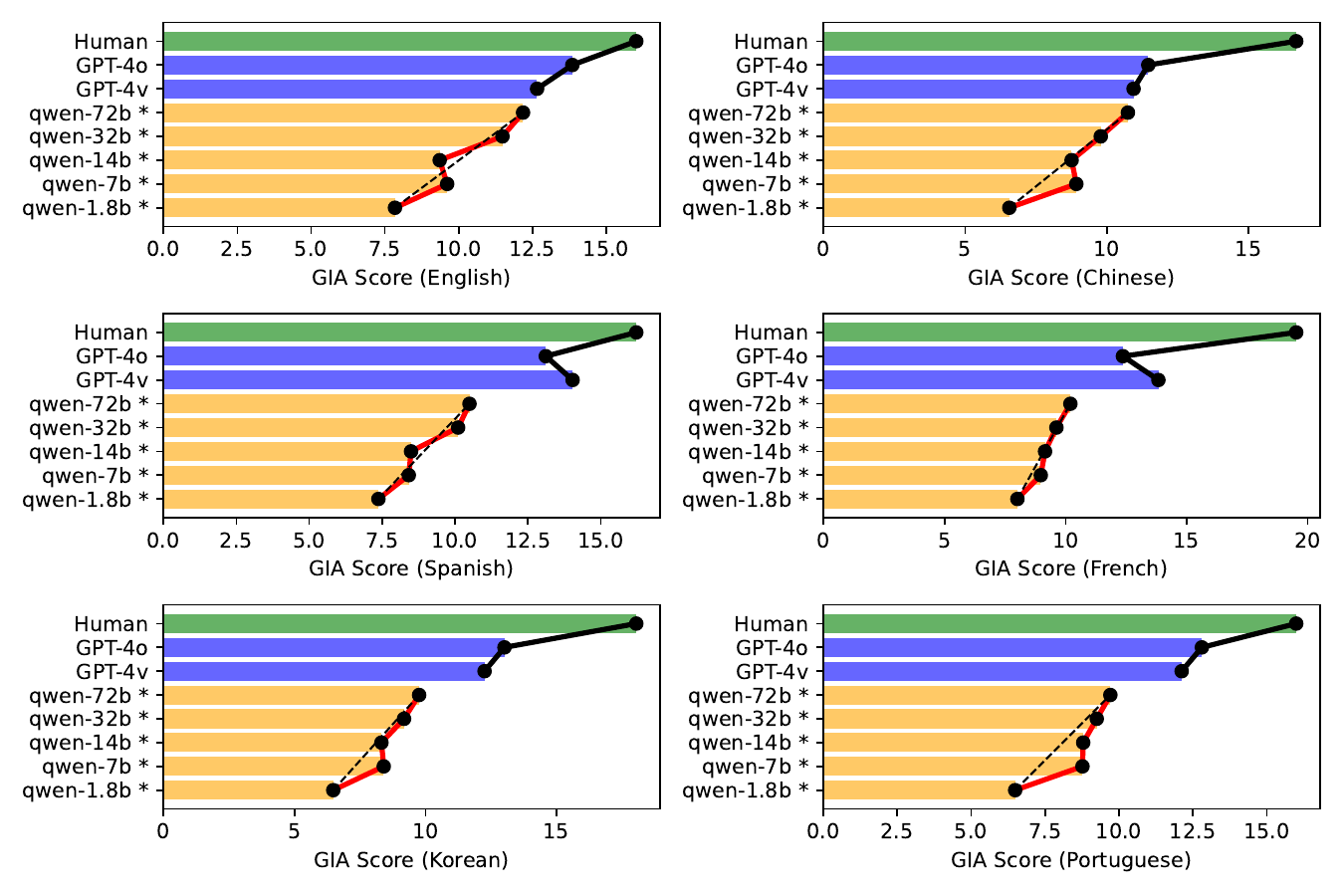}
  \caption{\textbf{The GIA scores across the six languages, with Qwen LLM series from 1.8B to 72B.} Generally, the GIA scores increase with the rise of LLM parameters. However, a threshold is observed when scaling up the LLMs' size from 7B to 14B.}
  \label{gia_fig}
\end{figure}

To further investigate the influence of LLM size to the GIA score, we conducted an ablation study with the Qwen series from 1.8B to 72B. In this experiment, we applied the LLaVA architecture and used the same ViT component (CLIP-ViT-L-14). In order to strictly control variables, we trained the models by ourselves using the same training data and the same set of hyperparameters for pretraining and fine-tuning. The data for pretraining is completely from LLaVA-1.5, and the data for fine-tuning is composed of LLaVA1.5~\cite{liu2023improved} dataset, ShareGPT4v~\cite{chen2023sharegpt4v} dataset and our private visual-text instruct data. We show the training data and hyperparameters for both first-stage vision-language alignment pretraining and the second-stage visual instruction tuning in Table.~\ref{tab:hyperparameters}. We use greedy decoding for evaluation to ensure reproducibility. The GIA scores on six languages are shown in Fig.~\ref{gia_fig}. 

Across the six languages analyzed, we consistently observe a significant increase in GIA scores with the expansion of LLM parameters. However, it is notably surprising that scaling up the size of LLMs from 7B to 14B parameters often yields no observable performance enhancement (and there might even be a slight decline).
This phenomenon suggests the existence of a threshold - indicative of an emerging point of general intelligence for MLLMs somewhere between 13B and 32B parameters. In other words, it indicates a potential threshold for attaining a superior level of general intelligence, likely situated in the parameter range of 13B to 32B.

\section{Case Study}
\label{case}

\begin{figure}[t]
  \centering
  \includegraphics[width=\textwidth]{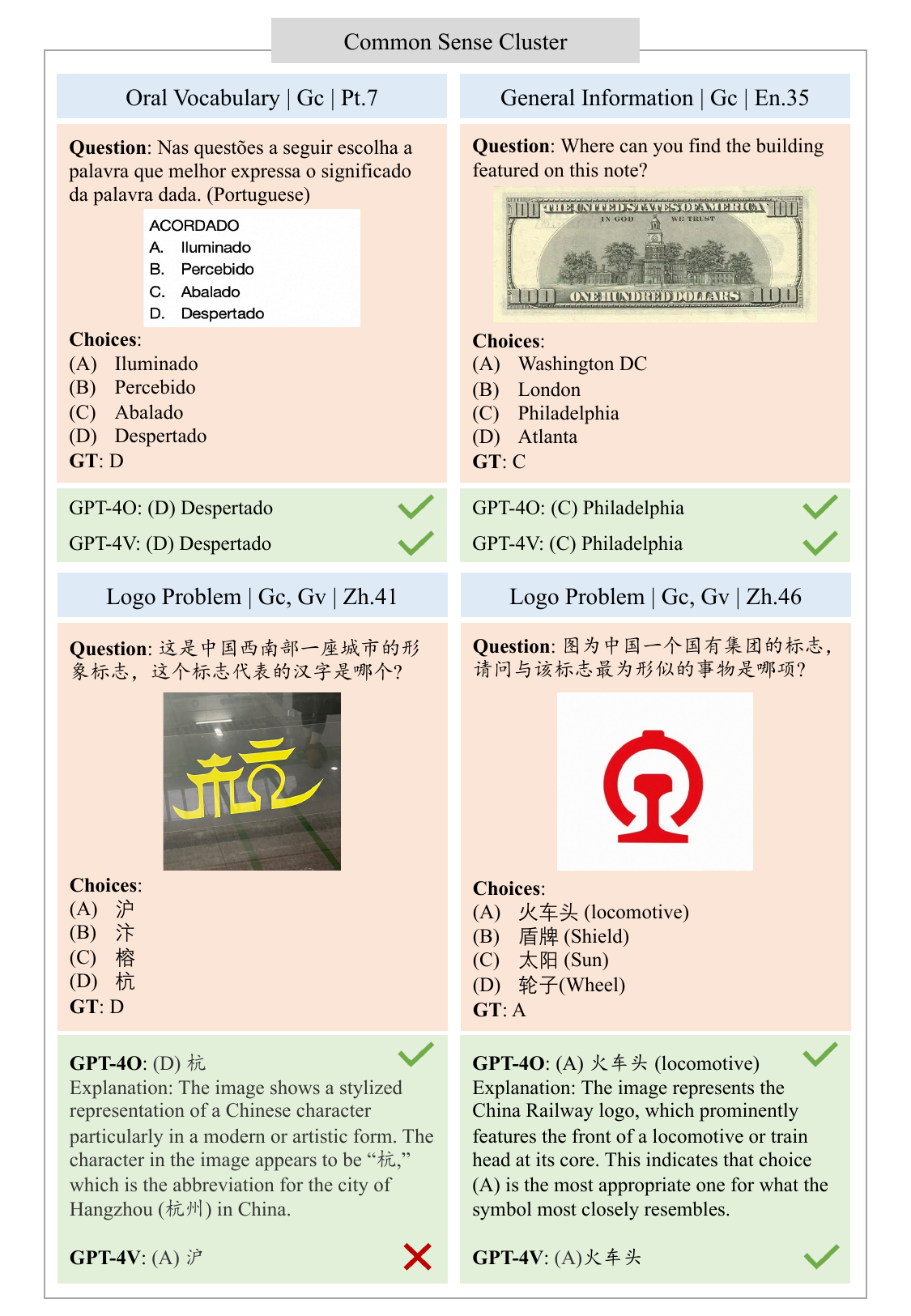}
  \caption{\textbf{Case Study of Common Sense Cluster.}}
  \label{common_sense_case_study_fig}
\end{figure}

\begin{figure}[t]
  \centering
  \includegraphics[width=\textwidth]{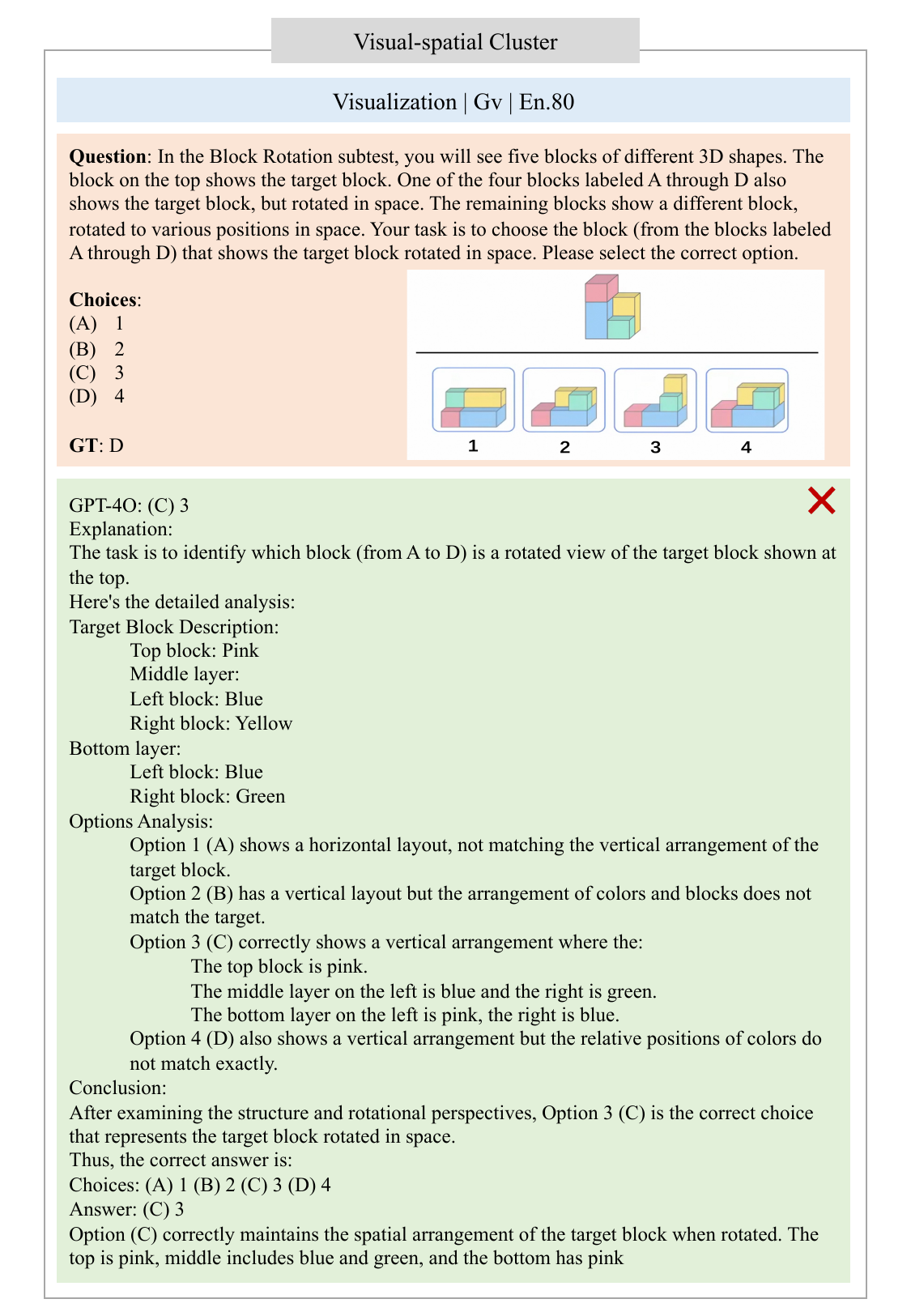}
  \caption{\textbf{Case Study of Visualization problem in Visual-Spatial Cluster.}}
  \label{visual_case_study_1_fig}
\end{figure}

\begin{figure}[t]
  \centering
  \includegraphics[width=\textwidth]{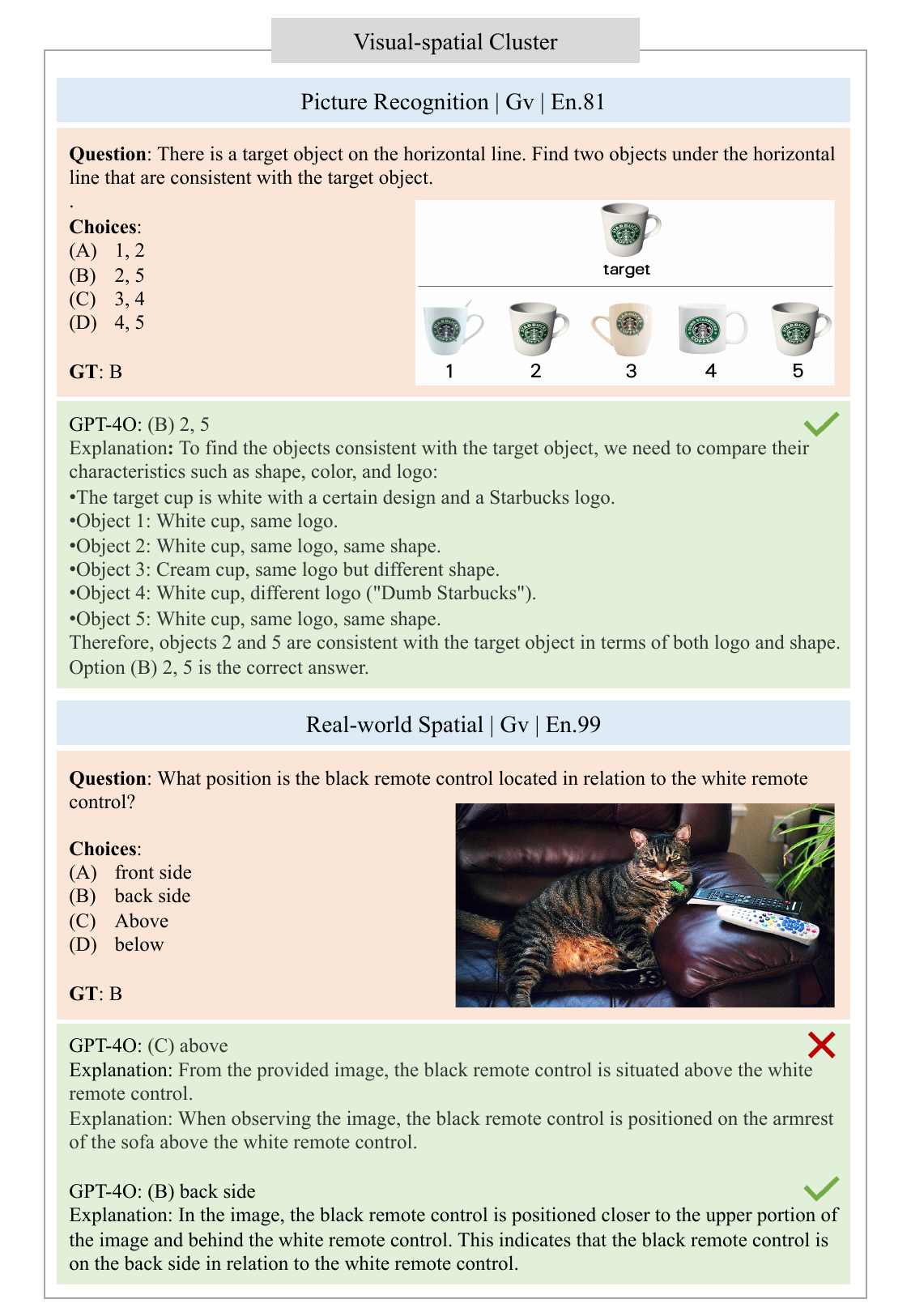}
  \caption{\textbf{Case Study of Picture Recognition and Real-world Spatial problem in Visual-Spatial Cluster.}}
  \label{visual_case_study_2_fig}
\end{figure}

\begin{figure}[t]
  \centering
  \includegraphics[width=\textwidth]{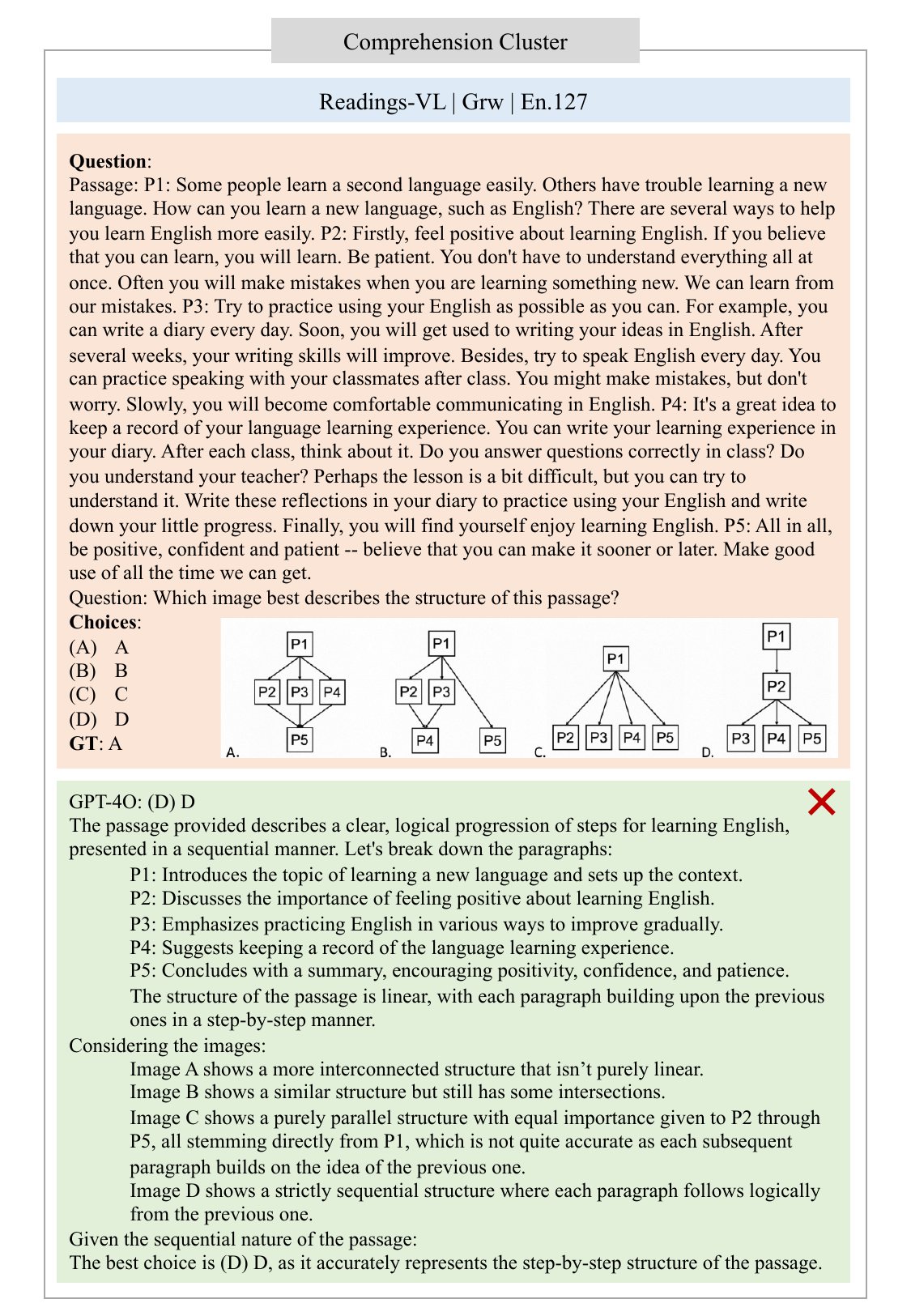}
  \caption{\textbf{Case Study of Readings-VL in Comprehension Cluster.}}
  \label{reading_case_study_1_fig}
\end{figure}

\begin{figure}[t]
  \centering
  \includegraphics[width=\textwidth]{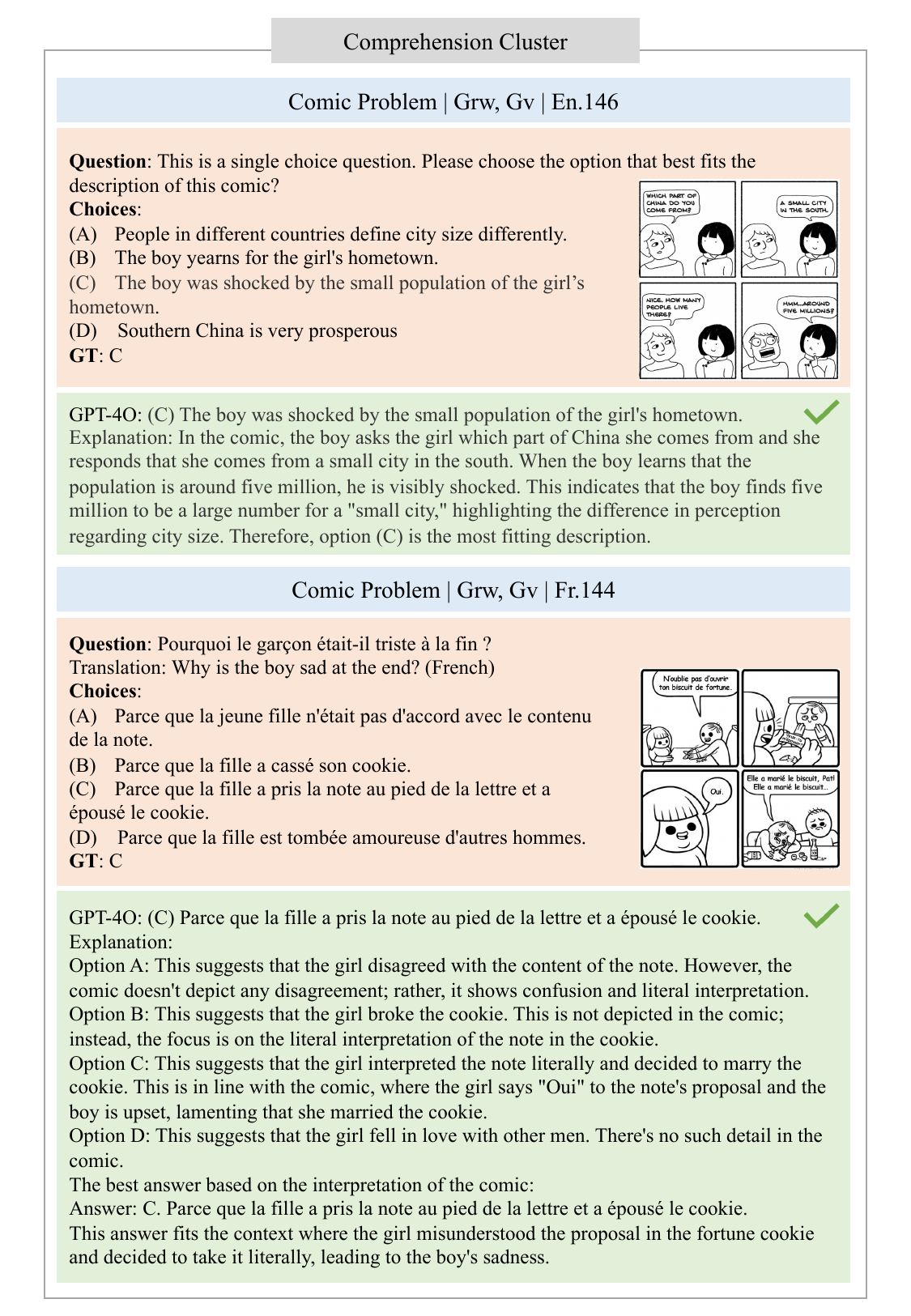}
  \caption{\textbf{Case Study of Comic Problem in Comprehension Cluster.}}
  \label{reading_case_study_2_fig}
\end{figure}

\begin{figure}[t]
  \centering
  \includegraphics[width=\textwidth]{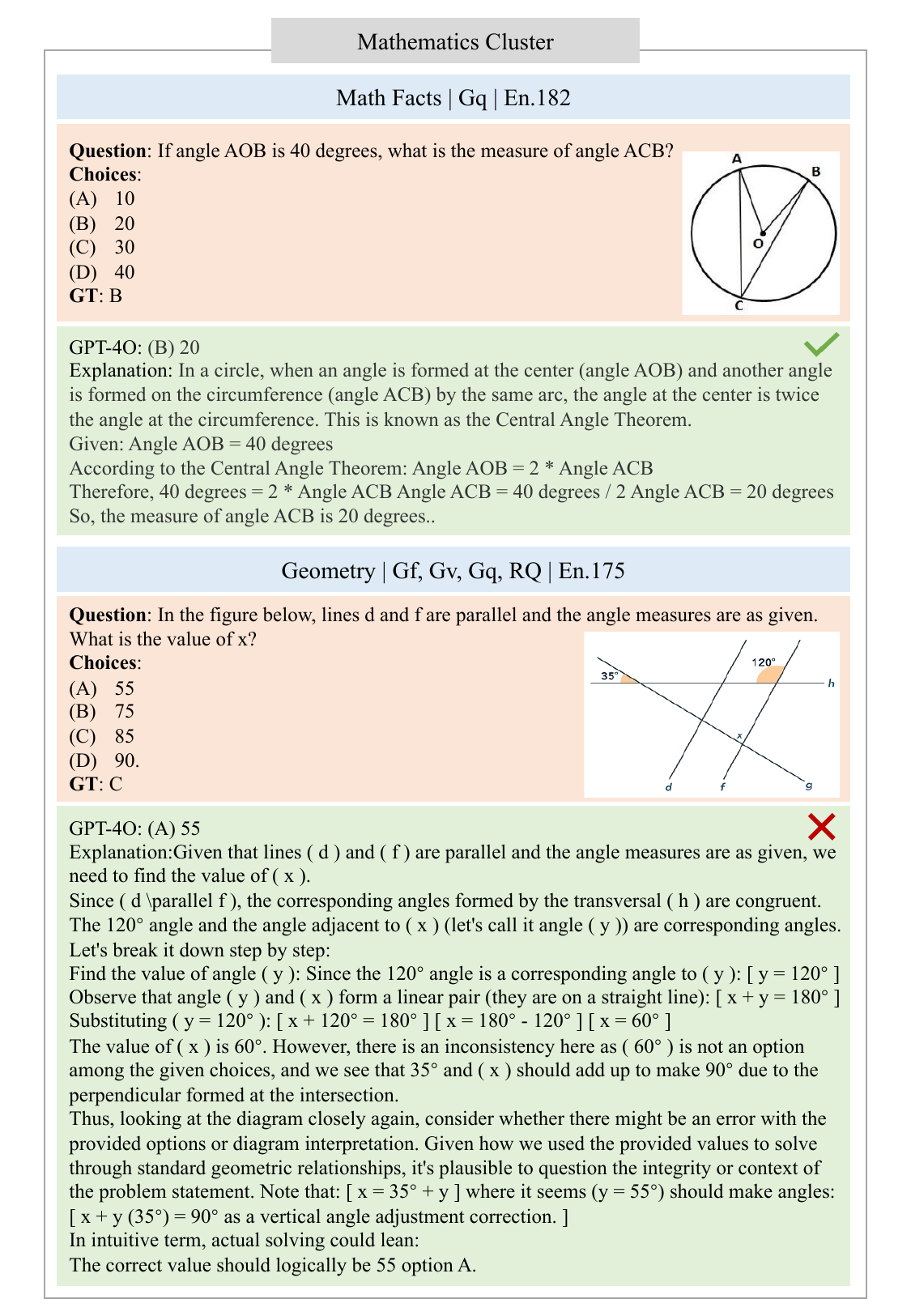}
  \caption{\textbf{Case Study of Readings-VL in Mathematics Cluster.}}
  \label{math_case_study_1_fig}
\end{figure}

\begin{figure}[t]
  \centering
  \includegraphics[width=\textwidth]{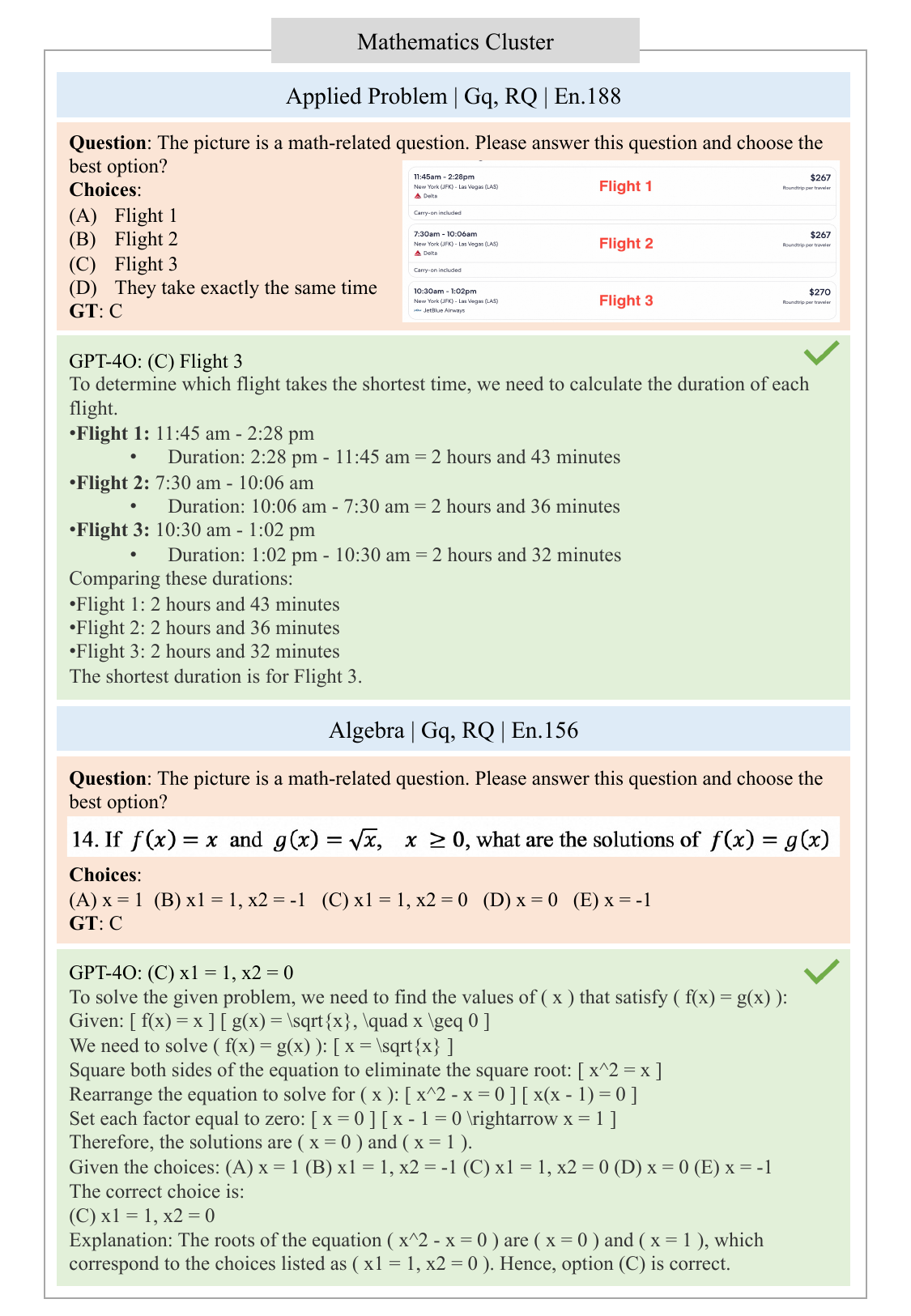}
  \caption{\textbf{Case Study of Comic Problem in Mathematics Cluster.}}
  \label{math_case_study_2_fig}
\end{figure}

\begin{figure}[t]
  \centering
  \includegraphics[width=\textwidth]{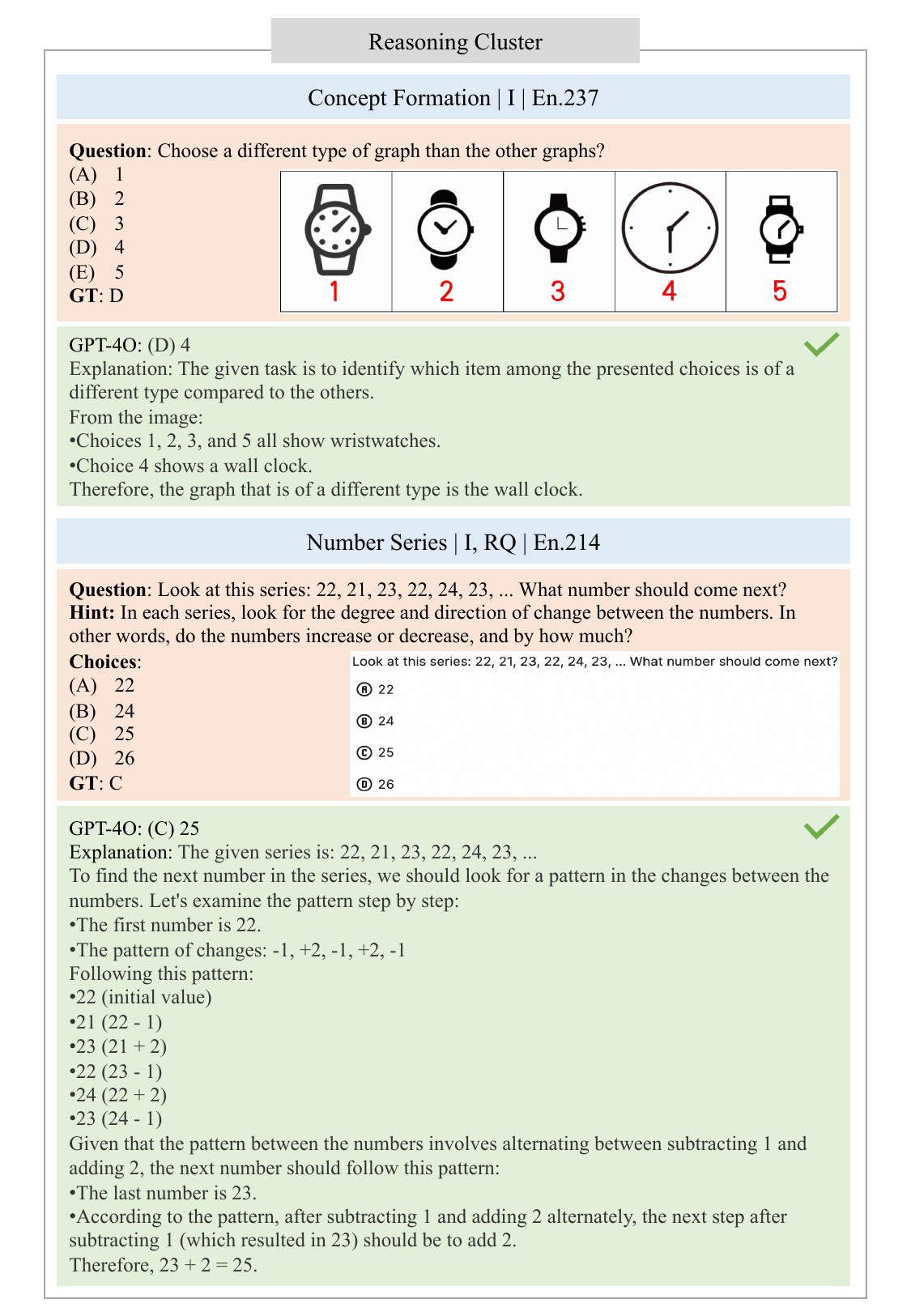}
  \caption{\textbf{Case Study of Number Series and Concept Formation problem in Reasoning Cluster.}}
  \label{reason_case_study_1_fig}
\end{figure}

\begin{figure}[t]
  \centering
  \includegraphics[width=\textwidth]{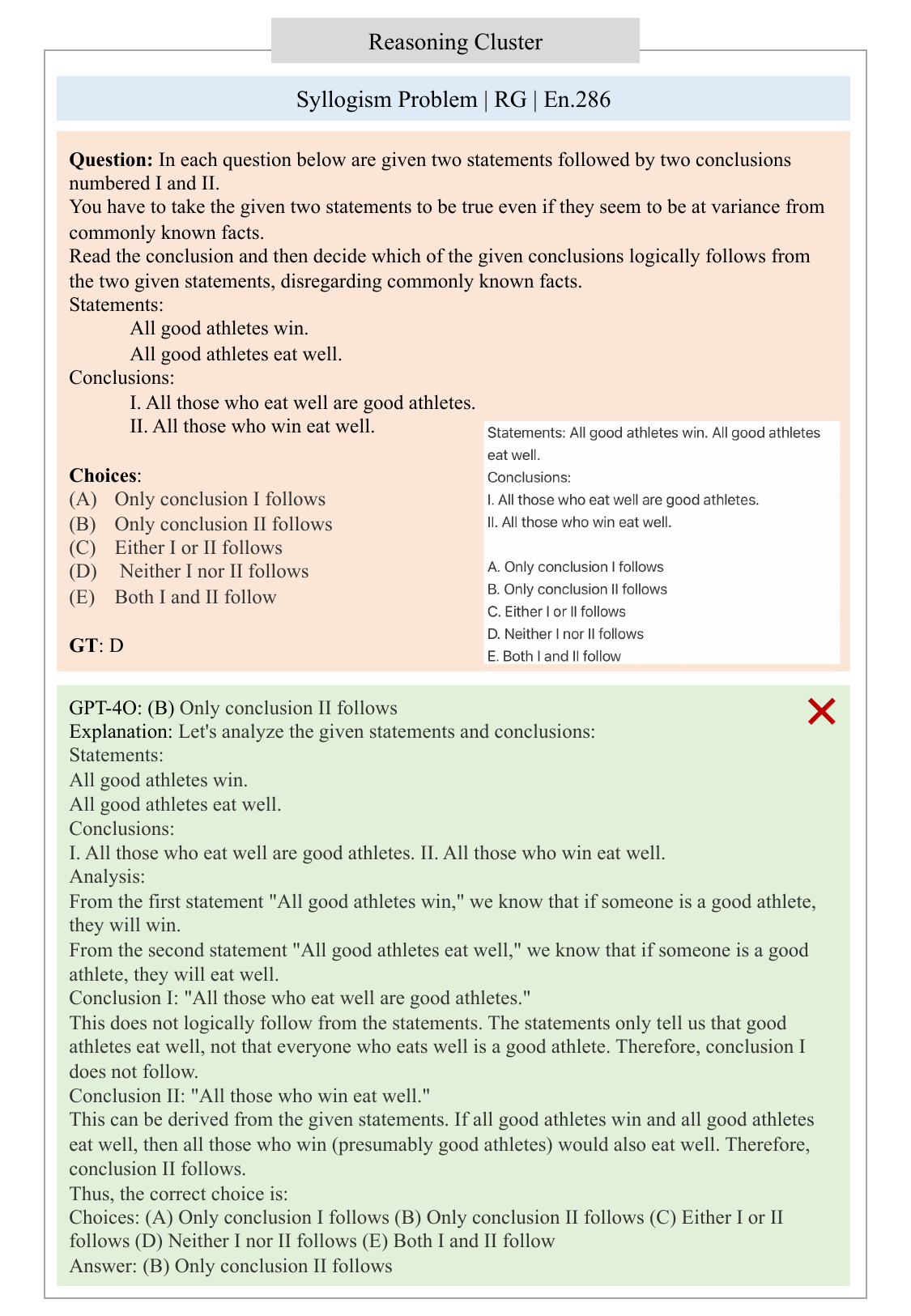}
  \caption{\textbf{Case Study of Syllogism Problem in Reasoning Cluster.}}
  \label{reason_case_study_2_fig}
\end{figure}

\begin{figure}[t]
  \centering
  \includegraphics[width=\textwidth]{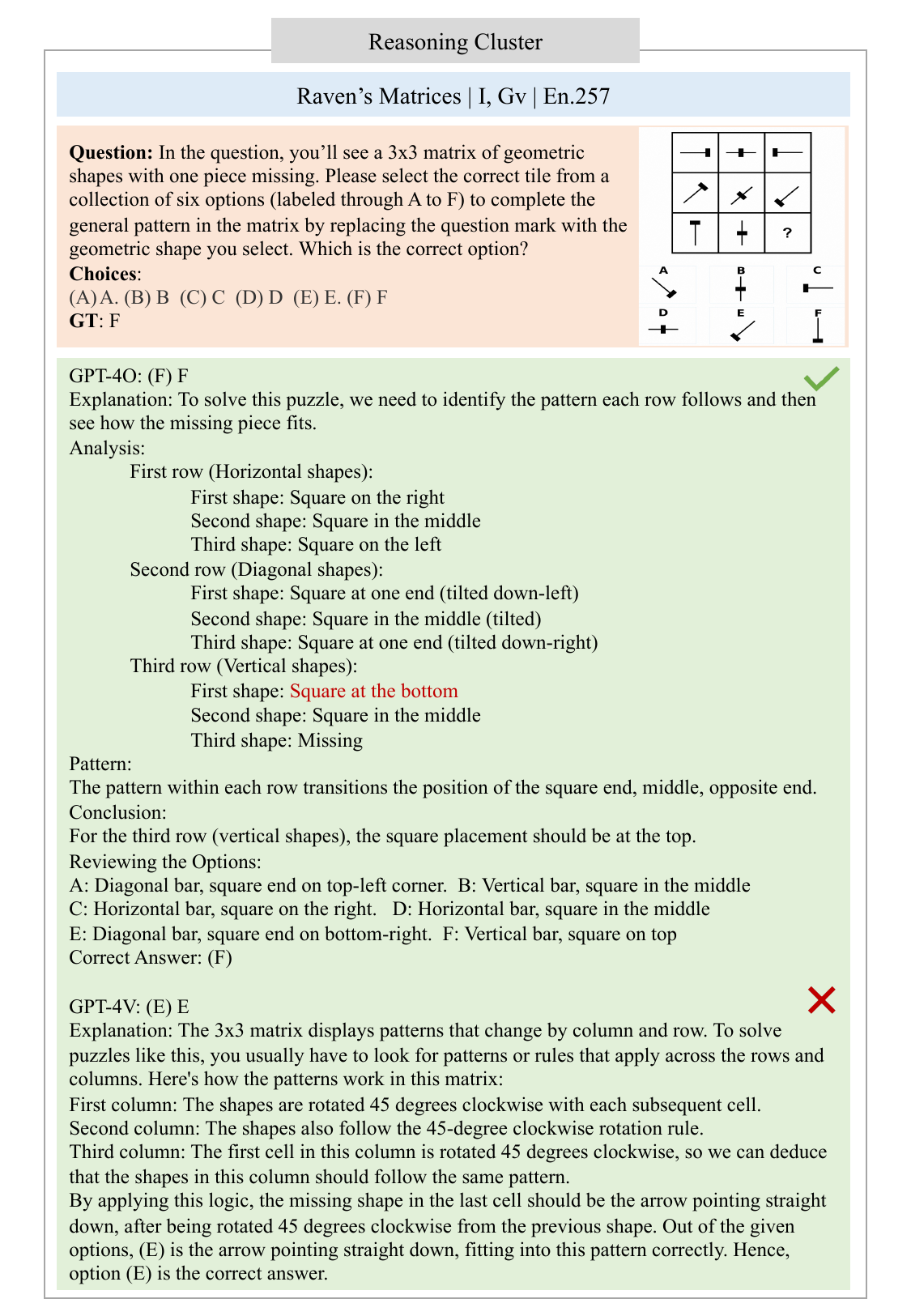}
  \caption{\textbf{Case Study of Raven’s Matrices Problem in Reasoning Cluster.}}
  \label{reason_case_study_3_fig}
\end{figure}

\subsection{The Common Sense Cluster}
\label{common_sense_cluster}

Current advanced MLLMs excel in common sense cluster, especially in general information and oral vocabulary questions, which can likely be bolstered by their extensive training datasets. However, there are still some deficiencies in logo problem related to cultural background for some MLLMs, e.g. GPT-4v. Logo problems usually contain confusing artistic characters or symbolic expression of cultural elements, which requires a high level of Gc and a certain level of Gv. As shown in Fig.~\ref{common_sense_case_study_fig}, GPT-4v can recognize the locomotive in the logo of chinese question 46, but it fails to recognize the Chinese character ("hang" in pinyin) in chinese question 41, while GPT-4o can perfectly recognize characters containing Chinese cultural elements.

\subsection{The Visual-spatial Cluster}
\label{visual_spatial_cluster}

In the Visual-spatial Cluster, current advanced MLLMs performe very well on the Picture Recognition questions, followed by the Real-world Spatial questions, and performed the worst on the Visualization transformation questions. The high accuracy on the Picture Recognition questions shows that the advanced MLLMs already has a good object recognition ability. Compared with object recognition ability, their ability to recognize three-dimensional spatial relationships is much worse, which can be divided into translation transformation and rotation transformation. The performance on the Real-world Spatial questions proves that the MLLMs can recognize the translation transformation relationship of objects in three-dimensional space with a certain probability, including up, down, left, right, front, and back. At the same time, the MLLMs suffer from the rotation transformation ability and spatial imagination ability in three-dimensional space, resulting the low accuracy on the Visualization transformation questions. As shown in Fig.~\ref{visual_case_study_2_fig}, after multiple inferences, GPT-4o can always recognize the same cup in english question 81 and the spatial relationship between the two remote controls with a high probability in english question 99, but it is difficult to recognize the same blocks after rotation in english question 80 in Fig.~\ref{visual_case_study_1_fig}.

\subsection{The Comprehension Cluster}
\label{comprehension_cluster}

Similar to the common sense cluster, current advanced MLLMs perform very well in comprehension cluster, including readings-text, readings-VL and the comic problem, which can be attributed to the powerful language capabilities of LLM. Surprisingly, GPT-4o understands the scenarios represented both visually and textually in comics quite well, which proves it can integrate visual narrative comprehension with textual comprehension. As shown in Fig.~\ref{reading_case_study_2_fig}, in english question 146 and french question 144, GPT-4o can understand the entire story's connotation based on the textual dialogues between characters and the plot development, especially can recognize the facial expressions and quantitative contrast of population in english question 146. At the same time, GPT-4o still has some shortcomings in understanding the relationship between text paragraphs. As shown in Fig.~\ref{reading_case_study_1_fig}, in english question 7, GPT-4o fails to capture the "general-specific-general" structure of the article.

\subsection{The Mathematics Cluster}
\label{mathematics_cluster}

This Mathematics cluster is designed to evaluate the Gq factor. Although current advanced MLLMs did not perform well on math problems overall, we found two interesting phenomena. One is that the model performs better on algebra problems than geometry problems, such as the english question 182 in Fig.~\ref{math_case_study_1_fig}. This may be attributed to the training data of LLM contains enough math knowledge text, but the visual module of MLLMs still has defects in abstract geometric figures and their relationships. The other is that the model performed better on math facts problems and problems that can be solved in one step by directly applying mathematical knowledge including symbolic knowledge and geometric knowledge than on problems that require multi-step reasoning. For example in Fig.~\ref{math_case_study_1_fig}, GPT-4o can apply the Central Angle Theorem to solve the english question 182, but fails to solve the english question 175 which needs multi-step reasoning and calculation. In addition, GPT-4o has reached a level of practical application in simple mathematical applied problem, such as the problem of choosing the shortest flight time in english question 188 as shown in Fig.~\ref{math_case_study_2_fig}.

\subsection{The Reasoning Cluster}
\label{reasoning_cluster}

The reasoning cluster is designed to evaluate the I (inductive) factor and RG (deductive reasoning) factor. Similar to the performance gap between geometry and algebra in mathematics cluster, there is also a performance gap between deductive and inductive reasoning. Although GPT-4o are approaching the average human level for deductive reasoning, it only marginally meet the passing line (60) on syllogism problem and real-world reasoning problem. For example in Fig.~\ref{reason_case_study_2_fig}, GPT-4o fails on the english question 286 which is a classic form of deductive reasoning and ask GPT-4o to decide which of the given conclusions logically follows from the two given statements. For inductive reasoning, GPT-4o performs quiet well in number series and concept formation problems, such as the english question 214 and 237 in Fig.~\ref{reason_case_study_1_fig}, but performs very poorly on the Raven's Matrices problems. Take the english question 254 in Fig.~\ref{reason_case_study_3_fig} as an example, GPT-4o mistakenly recognized the graphic in the third row and first column as a vertical line with a black square at the bottom, when it should actually be a black square at the top, resulting in the incorrect selection. GPT-4o can perform effective reasoning, but there is a certain probability that it will make small mistakes when recognizing graphics, which shows that its visual module needs to be further improved. In addition, we also show the results of GPT-4v, which misidentifies counterclockwise rotation as clockwise rotation and incorrectly identifies option E as the arrow pointing straight down. This proves GPT-4v is much worse than GPT-4o in both reasoning and visual recognition.

\section{Limitations}
\label{limitation}

\begin{itemize}[leftmargin=*]

    \item We have observed a phenomenon in MLLMs similar to human cognition known as ``winner takes all'', which corroborates the emergence of GIA within cutting-edge MLLMs. However, we have not yet been able to provide a more definitive and persuasive explanation for the underlying causes. Unraveling this will be one of the directions we dedicate ourselves to in the future.
    
    \item We have gathered human data to construct the GIA model and to compare the cognitive abilities of current MLLMs with those of humans. Yet, the human data we have amassed thus far is limited, which might impinge on the accuracy of the GIA model and the objectivity of our findings. Hence, we aim to continue maintaining and our dataset, as well as collecting more human participant data in the future to encompass a more comprehensive and varied set of human samples.
    
\end{itemize}

\bibliography{cite}
\bibliographystyle{neurips_data_2024}

\end{document}